%% file: icml2023.tex
\documentclass[nohyperref]{article}

\usepackage{microtype}
\usepackage{graphicx}
\usepackage{subfigure}
\usepackage{booktabs} 

\usepackage{hyperref}

\usepackage[accepted]{icml2023}

\usepackage{amsmath}
\usepackage{amssymb}
\usepackage{mathtools}
\usepackage{amsthm}
\usepackage{comment}
\usepackage[capitalize,noabbrev]{cleveref}

\theoremstyle{plain}
\newtheorem{theorem}{Theorem}[section]

\newtheorem{lemma}[theorem]{Lemma}

\theoremstyle{definition}
\newtheorem{definition}[theorem]{Definition}

\usepackage[textsize=tiny]{todonotes}

\icmltitlerunning{Leveraging Demonstrations to Improve Online Learning: Quality Matters}

\usepackage{smile}
\usepackage{extarrows}
\usepackage[OT1]{fontenc}
\usepackage{algorithm}
\usepackage{algorithmic}
\usepackage{scalerel}
\RequirePackage{natbib}
\usepackage{hyperref}
\hypersetup{
  pdftex,
  pdffitwindow=true,
  pdfstartview={FitH},
  pdfnewwindow=true,
  colorlinks,
  linktocpage=true,
  linkcolor=blue,
  urlcolor=blue,
  citecolor=blue
}
\usepackage[capitalize]{cleveref} 
\usepackage[textsize=tiny]{todonotes}

\newcommand{\ts}{\text{TS}}

\newcommand{\BR}{\mathfrak{BR}}

\renewcommand{\d}[1]{\operatorname{d}\!#1}

\input{macros}

\begin{document}

\twocolumn[
\icmltitle{Leveraging Demonstrations to Improve Online Learning: Quality Matters}



\icmlsetsymbol{equal}{*}

\begin{icmlauthorlist}
\icmlauthor{Botao Hao}{yyy}
\icmlauthor{Rahul Jain}{sch}
\icmlauthor{Tor Lattimore}{yyy}
\icmlauthor{Benjamin Van Roy}{yyy}
\icmlauthor{Zheng Wen}{yyy}
\end{icmlauthorlist}

\icmlaffiliation{yyy}{Deepmind}
\icmlaffiliation{sch}{University of Southern California, work done by Rahul Jain while at DeepMind}

\icmlcorrespondingauthor{Botao Hao}{haobotao000@gmail.com}
\icmlcorrespondingauthor{Rahul Jain}{rahuljai@usc.edu}

\icmlkeywords{Machine Learning, ICML}

\vskip 0.3in
]



\printAffiliationsAndNotice{\icmlEqualContribution} 

\begin{abstract}
We investigate the extent to which offline demonstration data can improve online learning. It is natural to expect some improvement, but \emph{the question is how, and by how much?} We show that the degree of improvement must depend on the \emph{quality} of the demonstration data. To generate portable insights, we focus on Thompson sampling (TS) applied to a multi-armed bandit as a prototypical online learning algorithm and model.  The demonstration data is generated by an expert with a given  \emph{competence} level, a notion we introduce.  We propose an informed TS algorithm that utilizes the demonstration data in a coherent way through Bayes' rule and derive a prior-dependent Bayesian regret bound. This offers insight into how pretraining can greatly improve online performance and 
how the degree of improvement increases with the expert's competence level. We also develop a practical, approximate informed TS algorithm through Bayesian bootstrapping and show substantial empirical regret reduction through experiments.
\end{abstract}

\input{intro}

\input{literature}

\input{setting}

\input{algorithm}

\input{theory}

\input{computation}
\input{empirical}

\input{conclusion}

\newpage
\bibliographystyle{icml2023} 
{\small
\bibliography{ref}
}

\input{appendix}
\end{document}

%% file: macros.tex
\newcount\Comments
\newcommand{\kibitz}[2]{\ifnum\Comments=1{\textcolor{#1}{\textsf{\footnotesize #2}}}\fi}

%% file: intro.tex
\section{Introduction}\label{sec:intro}

A modern paradigm for developing intelligent agents involves pretraining on large quantities of existing data followed by learning from real-time interactions.  For instance, to produce a chatbot, one can pretrain a large language model on text gathered from the internet and subsequently improve behavior through learning from interactions with humans \cite{ziegler2019fine, ouyang2022training}.  With such an approach, the preexisting text is treated as offline demonstration data that conditions a reinforcement learning agent before it engages in online learning.

It is natural to expect the offline demonstration data to improve performance of the online learning agent.  However, the degree of improvement must depend on the quality of the demonstration data.  If the data is produced by a \emph{competent} expert, it ought to improve the agent's performance more so than if not.  We study using the demonstration data to enhance the performance of an online learning agent in terms of regret minimization, formulating a notion of competence and an approach to learning from the demonstration data in a manner that accounts for this.

As a prototypical model for online learning, we consider multi-armed bandits that offer a simple context for understanding the role of offline data. We focus on Thompson sampling (TS) \cite{thompson1933likelihood} which is a popular online learning algorithm, owing to its
effectiveness across a wide range of environments and its scalability through the use of
approximation methods such as epistemic neural networks \citep{osband2021epistemic}. Moreover, TS offers a coherent way to use the demonstration data by simply following Bayes' rule. In our setting, pretraining amounts to conditioning the distribution used by TS as it initiates online learning.  
Note that a goal of this paper is to yield insights on how to leverage the demonstration data to improve online learning, and what the potential gains may be. Thus, we use the prototypical setting of a multi-armed bandit and a simple model for the data generation that is mathematically convenient and widely-used in machine learning.

\paragraph{Contributions} Our contribution is three-fold:

(i) We study how the quality of offline demonstration data can improve online learning and propose an \textit{informed} TS algorithm that naturally makes use of the offline data through an \emph{informative} prior. We show that the algorithm's online learning performance improves significantly with the quality of demonstrations as measured by a notion of the expert's competence level that we introduce.

(ii) We establish a \emph{prior-dependent} Bayesian regret bound that offers insight into how pretraining reduces regret and how this reduction depends on the expert's competence level. Previous works \cite{russo2016information, zhang2022feel, hao2022regret} mostly focus on \emph{prior-free} Bayesian regret bounds and 
thus cannot characterize the quality of offline data.
Our technique extends the information-theoretic regret decomposition to characterize how an informative prior can reduce both the information ratio and the entropy of the distribution of the optimal arm.

\noindent (iii) We propose a practical algorithm that approximates \emph{informed} TS via Bayesian bootstrapping. Through experiments, we show that a \textit{partially-informed} TS  algorithm that uses the offline demonstration data naively (i.e., assuming the expert to be naive in that it takes actions uniformly randomly) can only reduce the cumulative regret marginally. 
However, the \textit{informed} TS algorithm informed by the offline demonstration data and the competence level of the expert, can achieve substantial regret reduction.

%% file: literature.tex
\paragraph{Related Work.}
There is a rich body of literature on learning algorithms for bandits (see \citet{russo2018tutorial, lattimore2020bandit} for a detailed review).
Almost all of this literature assumes that the learning agent starts from scratch but this may lead to a long initial learning stage. In fact, offline data is available for many applications, such as training a large language model \cite{ouyang2022training}. 

There are some previous attempts that leverage offline data to warm-start an online learning algorithm. For instance, \citet{shivaswamy2012multi} analyzed a warm start UCB algorithm for the  $K$-armed bandit and 
\citet{zhang2019warm} investigated warm-starting contextual bandits by combining offline supervised feedback that is generated by an uniformly random policy. \citet{banerjee22artificial} proposed a meta-algorithm that uses historical data as needed to improve the computation and storage. However, none of these algorithms take the quality of offline data into consideration and thus show little regret reduction.

In the context of reinforcement learning (RL), there are several recent works \cite{rashidinejad2021bridging, xie2021policy, song2022hybrid, wagenmaker2022leveraging} that bridge offline and online RL. However, all of them focus on policy optimization rather than regret minimization. And they require different versions of concentrability coefficient conditions that are hard to be satisfied in practice. The importance of the competence level of the expert was first highlighted by \citet{beliaev2022imitation} for imitation learning and is modeled through an $\epsilon$-greedy policy. However, their goal is very different from ours since there is no online interaction there.

Offline data can also be viewed as a special form of side information and some other forms of side information are studied for online learning. \citet{degenne2018bandits} assumed the side information as observations of other arms while \citet{cutkosky2022leveraging} considered some hints about the optimal action for linear bandits.

%% file: setting.tex
\section{Problem Setting}
\label{sec:problem}

To offer a coherent formulation -- that is, one that conforms with standard axioms of statistical decision theory -- we model all unknown quantities as random variables defined with respect to a common probability space $(\Omega, \mathbb{F}, \mathbb{P})$. For a set $\cS$, we denote $|\cS|$ as its cardinality. For positive integer $N$, let $[N] := \{1,2, \ldots, N\}$. The $K \times K$ identity matrix is $\bI_K$.

We consider a stochastic $K$-armed linear bandit with action set $\mathcal{A}= \left \{a_1, a_2, \ldots, a_K \right \} \subseteq \mathbb R^d$. 
The environment is identified by a random vector $\theta \in \mathbb R^d$ with prior distribution $\nu_0(\cdot) = \mathbb{P}(\theta \in \cdot)$. The agent begins with an offline demonstration dataset $\cD_0 = \{(\bar{A}_n, \bar{R}_n)\}_{n=1}^N$ consisting of action-reward pairs. Then, at each time $t\in[T]$, the agent chooses an action $A_t\in\cA$ and receives a reward 
$$
R_t = \langle A_t, \theta\rangle + \eta_t\,,
$$
where $\langle \cdot , \cdot \rangle$ is the vector inner product and $(\eta_t)_{t=1}^T$ is a sequence of independent standard
Gaussian random variables. The experience thus far is recorded in the \textit{online} history $\cH_{t} = \{(A_\tau, R_\tau)\}_{\tau=1}^{t}$ with $\cD_t = (\cD_0, \cH_{t-1})$
denoting the entire dataset at time $t$. We write $\mathbb P_t(\cdot) = \mathbb P(\cdot|\cD_t)$ as the posterior measure where $\mathbb P$ is the probability measure over $\theta$ and the history and $\mathbb E_t(\cdot) = \mathbb E(\cdot|\cD_t)$.

A (learning) policy $\pi = (\pi_t)_{t\in\mathbb N}$ is a sequence of deterministic functions where $\pi_t(\cdot | \cD_t)$ specifies a probability distribution over $\cA$ conditioned on the dataset $\cD_t$. A stationary policy is an element of the probability simplex that does not depend on history. Let $A^*=\argmax_{a\in\cA} \theta^{\top} a$ and we define the Bayesian regret of a policy $\pi$ as 
\begin{equation*}
\BR_T(\pi) :=  \mathbb{E}\left[\sum_{t=1}^T \langle A^*, \theta\rangle - \sum_{t=1}^T R_t\right]\,,
\end{equation*}
where the expectation is over the environment $\theta$, the interaction sequence induced by the policy and environment 
and the offline demonstration data $\cD_0$.  

\subsection{Competence} 
Each action $\bar{A}_n$ in the offline demonstration dataset is generated by an expert and the expert's expertise level is characterized by a notion of {\it competence}. 

In particular, the expert's competence is parameterized by values $\lambda \geq 0$ and $\beta \geq 0$, which represent {\it knowledgeability} and {\it deliberateness}.  The expert's knowledge takes the form of a vector $\vartheta$, which is distributed as $\cN(\theta, \bI_d / \lambda^2)$ conditioned on $\theta$, and actions are selected according to an expert policy,
\begin{equation}
\mathbb{\phi}_{\beta, \lambda}(\bar{A}_n = a| \vartheta) =  \frac{\exp \left( \beta a^{\top}\vartheta \right)}{\sum_{b \in \mathcal{A}} \exp \left(\beta b^{\top}\vartheta \right)}\,.
\label{eq:softmaxactions}
\end{equation}
If $\lambda$ or $\beta$ is finite, the expert policy above takes suboptimal actions. One source of suboptimality stems from the agent's knowledgeability $\lambda$, 
which induces error in the agent's knowledge of $\theta$.  The other is due to deliberateness $\beta$, which drives the agent to choose actions that fail to optimize $\vartheta^\top a$. Together, this knowledgeability and deliberateness determine what we refer to as the agent's {\it competence}. 

The form of the expert policy is simple, mathematically convenient and widely used in reinforcement learning. The particular form we use is also expressive as we now note.

\begin{remark}
For multi-armed bandits where the actions are the basis vectors,  $\mathbb{\phi}_{\beta, \lambda}(\bar{A}_n = \cdot| \vartheta)$
is a random variable supported on the whole probability simplex.
In other words, if one views $\beta$ and $\vartheta$ as the parameters that parameterize the policy, the softmax policies with structure as in Eq.~\eqref{eq:softmaxactions} is enough to realize any stationary policy.

\end{remark}

\begin{remark}\label{remark_epsilon_greedy}
 \citet{beliaev2022imitation} consider an $\epsilon$-greedy expert policy (simplified from a MDP model): for $\beta\in[0, 1]$
\begin{equation*}
       \pi_\beta(\bar A_n=a|\theta) = \beta \mathbb I\left(a=\argmax_{a\in[K]}a^{\top}\theta\right) + (1-\beta)/K\,,
\end{equation*}
where $\mathbb I(\cdot)$ is the indicator function. However, they assume that the expert has perfect knowledge of the environment, e.g., the knowledgeability parameter $\lambda=\infty$, which is not realistic in practice.
\end{remark}

In general we expect that the insights developed for the specific model 
in \cref{eq:softmaxactions} will generalise to alternative models with similar qualitative properties.

%% file: algorithm.tex
\section{Informed Thompson Sampling}\label{sec:algorithm}

We introduce the  \emph{informed} Thompson Sampling (iTS) algorithm that uses the offline demonstration data in a coherent way. The details of the iTS algorithm are: 
\begin{enumerate}
    \item It constructs an \emph{informative} prior by the use of offline dataset $\cD_0$ and Bayes' rule, and which satisfies
\begin{equation}\label{eqn:posterior_update}
\begin{split}
     &\mathbb P(\theta\in\cdot|\mathcal{D}_0)
     \propto  \mathbb P(\mathcal{D}_0|\theta\in\cdot)\nu_0(\cdot)\\
    &\propto \nu_0(\cdot) \prod_{n=1}^{N}\underbrace{\mathbb P(\bar{A}_n | \theta \in \cdot)}_{\text{action likelihood}}\underbrace{\mathbb P(\bar R_{n}|\bar A_n, \theta\in\cdot)}_{\text{reward likelihood}}\,,
    \end{split}
\end{equation}
where $\nu_0(\cdot)$ is the initial prior for $\theta$.
\item The algorithm then uses the \emph{informative} prior to start learning and taking actions in the usual way: At time $t$, obtain a sample $\tilde{\theta}_t$ from the current posterior distribution $\mathbb P(\theta\in\cdot|\mathcal{D}_{t})$ and choose an arm $$A_t = \arg \max_{a \in \mathcal{A}} a^{\top}\tilde{\theta}_t\,.$$ Obtain reward $R_t$ and update the posterior distribution $\mathbb P(\theta\in\cdot|\mathcal{D}_t)$.
\end{enumerate}

Due to the action likelihood term, drawing samples from the exact posterior distribution $\mathbb P(\theta\in\cdot|\mathcal{D}_t)$ is hard even though we have a conjugate initial prior for $\theta$. In Section \ref{sec:approxsamples}, we propose an approximate-TS algorithm based on Bayesian bootstrapping. 

\begin{remark}
It is worth emphasizing here that the actions in the offline dataset also carry information about the environment through the action likelihood term $\mathbb P(\bar{A}_n | \theta \in \cdot)$ which can incorporate any information about the generative model of the policy used to generate the offline dataset, and thus greatly improve the informativeness of the prior.   
\end{remark}

%% file: theory.tex
\section{An Information-Theoretic Analysis}

We now present an information-theoretic regret analysis of the \textit{informed TS} algorithm. In particular, we demonstrate the role of the competence level though a \emph{prior-dependent} Bayesian regret bound.

In the literature, there are two ways to prove Bayesian regret bounds. The first is to introduce confidence sets such that the Bayesian regret bounds of TS match the best possible worst-case regret bounds of the UCB algorithm \citep{russo2014learning, zhang2022feel}. However, it is unclear how to use prior information or offline datasets to construct confidence sets in a principled way. There are some exceptions that attempt to show the effect of the prior distribution \citep{bubeck2013prior, kveton2021meta, simchowitz2021bayesian} but all rely on conjugate priors that do not hold for our setting.  

The second is to decompose the Bayesian regret into an \emph{information ratio} term and an \emph{entropy} term and bound them using tools from information theory \citep{russo2016information}. Next we show that existing analysis is not sufficient to fully characterize the prior effect.

\subsection{Why Existing Analysis Is Not Sufficient}

We reproduce the key step of existing information-theoretic analysis \citep{russo2016information}. Define the notion of information ratio:
\begin{equation*}
    \Gamma_t = \frac{\left(\mathbb E_t\left[\langle A^*, \theta\rangle-R_{t}\right]\right)^2}{\mathbb I_t(A^*; (A_t, R_{t}))}\,,
\end{equation*}
where $\mathbb I_t$ is the conditional mutual information\footnote{$\mathbb I_t(X; Y) = \mathbb E_t[D_{\text{KL}}(\mathbb P_{t, X|Y}||\mathbb P_{t, X})]$}.
\citet{russo2016information} bounded the Bayesian regret of the  TS algorithm as follows:
\begin{equation*}
   \BR_T(\pi^\ts)\leq \sqrt{\Gamma^*\cH(A^*)T}\,, 
\end{equation*}
where $\Gamma_t\leq \Gamma^*$ almost surely for any $t\in[T]$ and $\cH(\cdot)$ is the Shannon entropy. On the one hand, the  effect of the prior distribution can be partially characterized through its entropy. But this can only lead to logarithmic improvement since $\cH(A^*)$ is always bounded by $\log(K)$.

On the other hand, the upper bound on the  information ratio in \citet{russo2016information} is prior-independent. To see this,
using the probability matching property of the TS algorithm and Corollary 1 in \citet{russo2016information},
\begin{equation}\label{eqn:standard_bound}
    \begin{split}
  ( \mathbb E_t[\langle A^*, \theta\rangle&-R_{t}])^2
 = \left(\sum_{a\in\cA}\mathbb P_t(A^*=a)\Delta_t(a)\right)^2\\
  &\leq 2|\cA|\sum_{a\in\cA}\mathbb P_t(A^*=a)^2\Delta^2_t(a) \\
  &\leq |\cA|\mathbb I_t\left(A^*; (A_t, R_t)\right)\,,
    \end{split}
\end{equation}
where the first inequality follows from the Cauchy–Schwarz inequality and  $\Delta_t(a) = \mathbb E_t[\langle a, \theta\rangle|A^*=a]-\mathbb E_t[\langle a, \theta\rangle]$. However, the use of Cauchy–Schwarz inequality over the whole action set $\cA$ is agnostic to the distribution of $A^*$ and thus illuminates the effect of a prior. As far as we know, all the existing upper bound analysis of information ratio \citep{tossou2017thompson, lattimore2019information, hao2021information, hao2022regret} are prior-independent.

\subsection{A Novel Regret Decomposition}

We now introduce a novel information-theoretic regret decomposition such that the Bayesian regret bound can reflect the  effect of the prior distribution.  
Our proof template is general and can be used even when a more general form of the expert policy (than in Eq.~\eqref{eq:softmaxactions}) is used. Further, the proof technique can also be extended for other algorithms beyond TS, such as information-directed sampling \cite{russo2014learning}. 

\begin{definition}
Consider a random set $\cU\subseteq \cA$ which is measurable with respect to the offline dataset $\cD_0$. For any $0\leq\varepsilon\leq 1$, we call $\cU$ as \emph{$(1-\varepsilon)$-informative} if it contains the optimal action $A^*$ with probability at least $1-\varepsilon$:
\begin{equation}\label{def:informative_set}
   \mathbb P\left(A^*\in \cU\right)\geq 1-\varepsilon\,.
\end{equation}
\end{definition}
It is easy to see the full action set $\cA$ is $(1-\varepsilon)$-\emph{informative} for any $\varepsilon$. The goal is to find a $\cU$ whose expected cardinality is much smaller than $|\cA|$.

Given such a $\cU$, we can decompose the Bayesian regret based on whether $A^*$ belongs to $\cU$ or not:
\begin{equation}\label{eqn:regret_decom}
    \begin{split}
    \BR_T(\pi^{\ts})
    =& \mathbb E\left[\sum_{t=1}^T\sum_{a\in\cU}\mathbb P_t\left(A^*=a\right)\Delta_t(a)\right] \\
    &+\mathbb E\left[\sum_{t=1}^T\sum_{a\notin\cU}\mathbb P_t\left(A^*=a\right)\Delta_t(a) \right] \,.
    \end{split}
\end{equation}
Both terms can be bounded in terms of the expected cardinality of $\cU$ and $\varepsilon$ as we show in the following lemmas.
\begin{lemma}\label{lemma:decomp1}
Let $\cU$ be an $(1-\varepsilon)$-informative set defined in Eq.~\eqref{def:informative_set}. Then,  the following holds
\begin{equation}\label{eqn:upper_bound_infor}
\begin{split}
   & \mathbb E\left[\sum_{t=1}^T\sum_{a\in\cU}\mathbb P_t\left(A^*=a\right)\Delta_t(a)\right]\\
   &\leq \sqrt{T\mathbb E[|\cU|]\left(\log\left(\mathbb E[|\cU|]\right)+\varepsilon\log\left(K/\varepsilon\right)\right)}\,.
    \end{split}
\end{equation}
\end{lemma}
The proof is available in Appendix \ref{sec:proof_lemma1} and the key step is to use Cauchy–Schwarz inequality on a potentially much smaller set $\cU$ rather than $\cA$.  Eq.~\eqref{eqn:upper_bound_infor} sheds light on how the regret upper bound depends on $\mathbb E[|\cU|]$. In particular, $\mathbb E[|\cU|]$ is the upper bound for the \emph{information ratio} while $\log\left(\mathbb E[|\cU|]\right)+\varepsilon\log\left(K/\varepsilon\right)$ is the upper bound for the \emph{entropy}. When $|\cU|\ll|\cA|$, the upper bound on  the information ratio is much smaller than the bound in Eq.~\eqref{eqn:standard_bound}.

\begin{lemma}\label{lemma:decomp2}
Let $\cU$ be an $(1-\varepsilon)$-informative set defined in Eq.~\eqref{def:informative_set} and suppose the expected reward range $\mathbb E[\max a^{\top}\theta-\min a^{\top}\theta]$ is bounded by $C_1$. Then,  the following holds
\begin{equation*}
\begin{split}
   \mathbb E\left[\sum_{t=1}^T\sum_{a\notin\cU}\mathbb P_t\left(A^*=a\right)\Delta_t(a)\right]\leq C_1T\varepsilon\,.
    \end{split}
\end{equation*}
\end{lemma}
The proof is available in Appendix \ref{sec:proof_lemma2}. This is an additive term that captures the regret when the informative set fails to contain the optimal action. This term is always negligible since in most cases $\varepsilon$ decays exponentially fast. 

Combining Lemmas \ref{lemma:decomp1} and \ref{lemma:decomp2} together, we have the following theorem.
\begin{theorem}\label{thm:main_general}
For any $(1-\varepsilon)$-informative set $\cU$, the Bayesian regret of TS algorithm can be upper bounded as 
\begin{equation*}
    \begin{split}
       & \BR_T(\pi^{\ts})\\
       &\leq  \sqrt{T\mathbb E[|\cU|]\left(\log\left(\mathbb E[|\cU|]\right)+\varepsilon\log\left(K/\varepsilon\right)\right)}+ C_1T\varepsilon\,.
    \end{split}
\end{equation*}
\end{theorem}
If such a $\cU$ is given to the algorithm (not limited to TS) as a prior knowledge, we can easily achieve the bound in Theorem \ref{thm:main_general}, e.g., running standard UCB on $\cU$ directly. In contrast, TS does not need to know $\cU$ and can adapt to different $\cU$ automatically. Thus, for TS, the introduction of $\cU$ is for \emph{analysis only} rather than used by the algorithm. 
\begin{remark}
Of course, the Bayesian regret bound of TS cannot exceed $O(\sqrt{T|\cA|\log(|\cA|)})$ by using the standard prior-free analysis \cite{russo2016information}. 
\end{remark}

Next, we use this result to bound the regret for the iTS algorithm for Gaussian bandits by finding such an informative set.

\subsection{Prior-Dependent Regret Bound For Informed-TS}  

Consider a $K$-armed bandit and assume the prior distribution $\nu_0(\cdot) = \cN(0, \bI_K)$.
We define ${\cU}_A$ (that is the set $\cU$ we choose in Section 4.2) as a \emph{set} that contains non-duplicated actions appearing in $\{\bar A_1, \ldots, \bar A_N\}$ at least once and $\cU_A$ has at most $K$ different actions. 

Let us first denote $\alpha_1 = K\min\{\log(T\beta)/\beta, 1\}$ and 
       $\alpha_2 = \exp(\beta \sqrt{2\log(TK)}/\lambda)$.
We further denote
\begin{equation*}
\begin{split}
     & f_1 =  \frac{3}{T}+\left(1-\frac{1}{\alpha_2(1+\alpha_1+\frac{\log(T)}{\log(1/\alpha_1)}+K/(T\beta))}\right)^N\,,\\
     &f_2 = \min\left\{ \alpha_1+1+\alpha_2\frac{KN}{T\beta}+\frac{1}{T}, K\right\}\,.
\end{split}
\end{equation*}

\begin{lemma}\label{lemma:lemma1}
If $K\geq \log_2(T)$, then the set $\cU_A$ is $(1-f_1)$-informative and $
\mathbb E[|\cU_A|]\leq f_2\,.
$
\end{lemma}
The proof can be found  in Appendix \ref{sec:proof_lemma1}. Here, $\alpha_1$ is the price for deliberateness and goes to 0 as the deliberateness level $\beta$ increases. $\alpha_2$ is the price the agent pays for the imperfect knowledge of the true environment $\theta$ and goes to 1 as the knowledgeability $\lambda$ tends to infinity. $f_1$ is the probability that $\cU_A$ fails to capture the optimal action and decays exponentially fast as the data size $N$ increases.   
Note that $K\geq \log_2(T)$ is
 a technical condition that we hope to relax in the future.

Since $\theta\sim\mathcal{N}(0, \bI_K)$, we  have
\begin{equation}\label{eqn:reward_range}
    \mathbb E\left[\max_a\langle a,\theta\rangle-\min_a\langle a, \theta\rangle\right] \leq 2\sqrt{2\log(K)}\,,
\end{equation}
where the proof of this claim is available in Appendix \ref{sec:proof_gaussian_range}.
Combining Theorem \ref{thm:main_general}, Lemma \ref{lemma:lemma1} and Eq.~\eqref{eqn:reward_range} together, we obtain the final regret bound for the informed TS algorithm, the main theoretical result of this paper.
\begin{theorem}\label{thm:main-regretbound}
The Bayesian regret of the \textit{iTS} algorithm is bounded as
\begin{equation}\label{eqn:final_bound}
\begin{split}
     \BR_T(\pi^{\text{i-TS}})&\leq  \underbrace{2\sqrt{2\log(K)}Tf_1+4\sqrt{2\log(K)}}_{\text{remainder term}}\\
     &+\underbrace{\sqrt{Tf_2\left( \log(f_2) + f_1\log\left(K/f_1\right)\right)}}_{\text{main term}}\,.
\end{split}
\end{equation}
\end{theorem}
For the main term, as the deliberateness $\beta$ increases,
the information ratio part $(f_2)$ first drops polynomially until 1 and then the entropy part $( \log(f_2) + f_1\log\left(K/f_1\right))$ drops further until $\log(1)=0$. This implies we must sharpen both the information ratio term and entropy term.
Overall, as the competence parameters $\beta, \lambda$ of an expert go to infinity, the main term of the Bayesian regret bound goes to 0. Thus, our regret bound Eq.~\eqref{eqn:final_bound} can fully characterizes the role of competence.

%% file: computation.tex
\section{Bayesian Bootstrapping}\label{sec:approxsamples}
The posterior update in Eq.~\eqref{eqn:posterior_update} is computationally challenging due to the loss of conjugacy in the $\mathbb P(\bar{A}_n | \theta)$ term while using the Bayes' rule, which has a sum of exponentials term in the denominator. Thus, we adapt the existing approximate-TS approach based on \textit{Bayesian bootstrapping} \citep{osband2019deep, lu2017ensemble}. The key idea is to perturb the loss function for the maximum a posterior (MAP) estimate and use the point estimate as a surrogate for the exact posterior sample.

\subsection{The Loss Function and Perturbation}\label{sec:bootstrapping}
We now introduce a loss function whose optimization we show yields the MAP estimate of the model parameters.  Suppose $\nu_0(\cdot)=\cN(0, \Sigma_0)$. With a bit of notational ambiguity, we view $\theta$ and $\vartheta$ as parameters rather than random variables in this subsection. We first derive the loss function for the MAP estimate in Lemma \ref{lemma:MAP}. 

\begin{lemma}\label{lemma:MAP}
At time $t$, the MAP estimate for $\theta, \vartheta$ is equivalent to solving the following optimization problem:
\begin{equation*}
    \begin{split}
   \argmax_{\theta, \vartheta}\cL_1(\theta, \vartheta) + \cL_2(\theta, \vartheta)+\cL_3(\theta, \vartheta)\,,
    \end{split}
\end{equation*}
where $\cL_1(\theta, \vartheta):=$
\begin{equation*}
\begin{split}
&- 2\sum_{n=1}^{N}\left(\beta \vartheta^T \bar{A}_n - \log\left( \sum_{b \in \mathcal{A}} \exp \left( \beta \vartheta^T b \right)\right)\right)\,,
\end{split}
\end{equation*}
is the negative log-likelihood contributed by the offline actions and $\cL_2(\theta, \vartheta):=$
\begin{equation*}
\begin{split}
 \sum_{n=1}^{N} \left( \bar R_{n} - \theta^T \bar{A}_n \right)^2 + \sum_{\tau=1}^{t} \left( R_{\tau} - \theta^T A_{\tau} \right)^2\,,
\end{split}
\end{equation*}
is the negative log-likelihood contributed by the rewards and $\cL_3(\theta, \vartheta):=$
\begin{equation*}
\begin{split}
\lambda^2\| \vartheta-\theta\|_2^2+\theta^{\top}\Sigma_0^{-1}\theta\,,
\end{split}
\end{equation*}
is  the log-prior function.
\end{lemma}
The proof is available in Appendix \ref{sec:proof_map}. Note that $\cL_1(\theta, \vartheta)$ serves a role similar to the imitation learning loss \citep{ross2011reduction} since it regularizes the online agent to follow the expert's action and the amount of regularization is guided by the competence level. While there are multiple heuristic choices for the imitation learning loss in literature, ours are derived in a principled way following Bayes' rule and combined with online learning.

\begin{remark}
When $\lambda$ is small, the competence level of the expert is low such that the demonstration data is of low-quality. In this case, the online learning agent should not be following the offline action. Fortunately, informed TS can understand this automatically through Bayes' rule. In particular, small $\lambda$ imposes little regularization on $\| \vartheta-\theta\|_2$ such that estimating $\theta$ is independent of the action likelihood.
\end{remark}

Now, we will perturb the loss function. As is standard in the literature \cite{lu2017ensemble, osband2018randomized, dwaracherla2022ensembles, qin2022analysis}, the Gaussian rewards are perturbed by additive Gaussian noise while the  offline actions are perturbed by multiplicative random weights. Moreover, the log-prior terms are perturbed by random samples from the prior distribution. Therefore, we denote a set of perturbations as follows and resample all the perturbations at each time:
\begin{itemize}
    \item \emph{Action perturbation.} Let $w_n\overset{\text{i.i.d}}{\sim} \exp(1)$ be a sequence of Bayesian bootstrap weights. The corresponding perturbed loss function is $\tilde \cL_1(\theta, \vartheta):=$
    \begin{equation*}
    \begin{split}-2\sum_{n=1}^{N}w_n\left(\beta \vartheta^T \bar{A}_n - \log\left( \sum_{b \in \mathcal{A}} \exp \left( \beta \vartheta^T b \right)\right)\right)\,.
        \end{split}
    \end{equation*}
    \item \emph{Reward perturbation.} Let $\xi_n^1, \xi_{\tau}^2\overset{\text{i.i.d}}{\sim} \cN(0, 1)$. The corresponding perturbed loss function is $ \tilde \cL_2(\theta, \vartheta):=$
    \begin{equation*}
        \begin{split}
        \sum_{n=1}^{N} \left(\bar R_{n}+\xi^1_n - \theta^T \bar{A}_n \right)^2+ \sum_{\tau=1}^{t} \left( R_{\tau}
        +\xi^2_{\tau}  - \theta^T A_{\tau} \right)^2\,.
        \end{split}
    \end{equation*}
    \item \emph{Prior function perturbation.} Let $\tilde \theta_0 \sim \cN(0, \Sigma_0)$ and $\tilde  \vartheta\sim \cN(0,  \bI_d/\lambda)$. The corresponding perturbed loss function is $ \tilde \cL_3(\theta, \vartheta):= $
    \begin{equation*}
      \lambda^2\|\vartheta-\theta-\tilde \vartheta\|_2^2+(\theta - \tilde \theta_0)^{\top}\Sigma_0^{-1}(\theta-\tilde \theta_0)\,.
    \end{equation*}
\end{itemize}

At each time $t$, let $(\hat{\theta},\hat{\vartheta})$ be the solution of 
\begin{equation}\label{eqn:perturb_loss}
    \min_{\theta, \vartheta} \tilde\cL_1(\theta, \vartheta) + \tilde\cL_2(\theta, \vartheta)+\tilde\cL_3(\theta, \vartheta)\,,
\end{equation}
and use $\hat{\theta}$ as a surrogate for posterior sampling in the standard TS algorithm. Since the perturbed loss function in Eq.~\eqref{eqn:perturb_loss} is convex, we can solve it by use of standard convex optimization solvers such as CVXPY \citep{diamond2016cvxpy}. 

\begin{remark}
1. We would like to mention perturbation-based methods such as Bayesian bootstrapping lead to one kind of approximate-TS algorithms. There are  other choices such as MCMC, Laplace approximation or variational inference for sampling from an approximate posterior. For a detailed empirical comparison, we refer the reader to \citet{osband2022the}.

2. While the algorithm as presented above assumes the offline dataset comes from a single expert, it can easily be extended where it comes from multiple experts with different competence parameters $(\beta_j,\lambda_j), j=1,\cdots, J$, with corresponding ($\vartheta_j$) parameters, and dataset sizes $N_j$. Namely, there will be $J$ similar terms in the loss function  $\tilde \cL_1$, one for each expert. Similarly, the first term in the loss function $\tilde \cL_3$ will be replaced by $J$ identical terms, one for each expert.
\end{remark}

\subsection{Estimating Competence Level}\label{sec:est_competence}
Bayesian bootstrapping requires an input for the competence level. In practice, this is often not available and can  be estimated only from the offline data. We provide two methods to estimate the deliberateness parameter, $\beta$:

\noindent 1) The first method is based on maximum likelihood estimation (MLE). Similar idea has been proposed to estimate the expertise level in imitation learning \cite{beliaev2022imitation}. Specifically, we optimize $\beta$ over the following negative log-likelihood of the offline data:
\begin{equation*}
\begin{split}
-\sum_{n=1}^{N}\left(\beta \bar{A}_n^{\top}\hat\vartheta^{\text{LS}}  - \log\left( \sum_{b \in \mathcal{A}} \exp \left( \beta b^{\top}\hat\vartheta^{\text{LS}} \right)\right)\right)\,,
\end{split}
\end{equation*}
where $\hat\vartheta^{\text{LS}}$ is the regularized least square estimate using $\cD_0$.

\noindent 2) The second method is to simply look at the entropy of the empirical distribution of the action in the offline dataset. Suppose the empirical distribution of $\{\bar A_n\}_{n=1}^N$ is $\mu_A$. Then we use $c_0/\cH(\mu_A)$ as an estimation for $\beta$, where $c_0>0$ is a hyperparameter. The intuition is that for smaller $\beta$, the offline actions tend to be more uniform and thus the entropy will be large. This is an unsupervised approach and agnostic to specific offline data generation process. 

The knowledgeability $\lambda$ is not quite `estimable'  because for a single environment,  even though we know the true environment $\theta$ and the expert's knowledge $\vartheta$, we only have one pair of observations. Thus,  the variance of the estimation for $\lambda$ could be infinite. However, exact estimation of $\lambda$ is not often necessary and we show that our algorithm is robust to misspecified $\lambda$ through experiments in Section \ref{sec:experiment_mis}.

We summarize the full (Bayesian bootstrapped) approximate iTS algorithm in Algorithm \ref{alg:uncertainty}.
\begin{algorithm}[ht!]
\caption{Approximate iTS} 
\begin{algorithmic}[1]
\label{alg:uncertainty}
\STATE \textbf{Input:} time horizon $T$, action set $\cA$, parameter $\lambda_0$, offline demonstration data $\mathcal{D}_{0}$;

\STATE Obtain $\hat{\beta}$ through either MLE or entropy method described in Section \ref{sec:est_competence}.

\FOR{$t=1, \ldots, T$}
\STATE Sample a set of perturbations $\{w_n, \xi_n^1, \xi_{\tau}^2,\tilde \theta_0 ,\tilde \vartheta\}$ according to Section \ref{sec:bootstrapping}.
\STATE Solve Eq.~\eqref{eqn:perturb_loss} with competence level $(\hat{\beta}, \lambda_0)$ and denote the solution as $(\hat{\theta}_t,\hat{\vartheta}_t)$.
\STATE Take action $A_t = \argmax_{a\in\cA} a^{\top}\hat{\theta}_t$ and receive $R_t$.
\ENDFOR
\end{algorithmic}
\end{algorithm}

%% file: empirical.tex
\section{Empirical Results}\label{sec:empirical}
We empirically investigate the role of offline demonstration data in terms of regret reduction. We compare the (approximate) \textit{informed} TS algorithm with two baseline algorithms: 

\noindent$\circ$ \textit{Uninformed} TS: An algorithm that uses the standard linear Gaussian TS \cite{russo2018tutorial} and does not use the offline demonstration data $\cD_0$;

\noindent $\circ$  \textit{Partially-informed} TS: An algorithm that uses the offline demonstration data $\cD_0$ to update the initial prior but assumes the expert is \textit{naive}, i.e., $\beta=0$ and hence the action likelihood term $\mathbb P(\bar{A}_n | \theta \in \cdot)$ is a constant, e.g.,
\begin{equation*}
\begin{split}
     \mathbb P(\theta\in\cdot|\mathcal{D}_0)
    \propto \nu_0(\cdot) \prod_{n=1}^{N}\mathbb P(\bar R_{n}|\bar A_n, \theta\in\cdot)\,.
    \end{split}
\end{equation*}
Note that this algorithm can use conjugacy in updating the posterior distribution, and hence is computationally simpler.

The environment draws a model $\theta$ from the prior distribution $\cN(0, \bI_d)$ and provides a noisy $\vartheta\sim \cN(\theta, \bI_d/\lambda^2 )$ to the expert. The expert then generates demonstration data following Eq.~\eqref{eq:softmaxactions} to obtain a dataset of size $N$. Each algorithm is run for a horizon $T=1000$ and we compute the average cumulative regret over 100 independent runs for each algorithm.

\subsection{Role of Competence}
\begin{figure}[t!]
\includegraphics[width=0.23\textwidth]{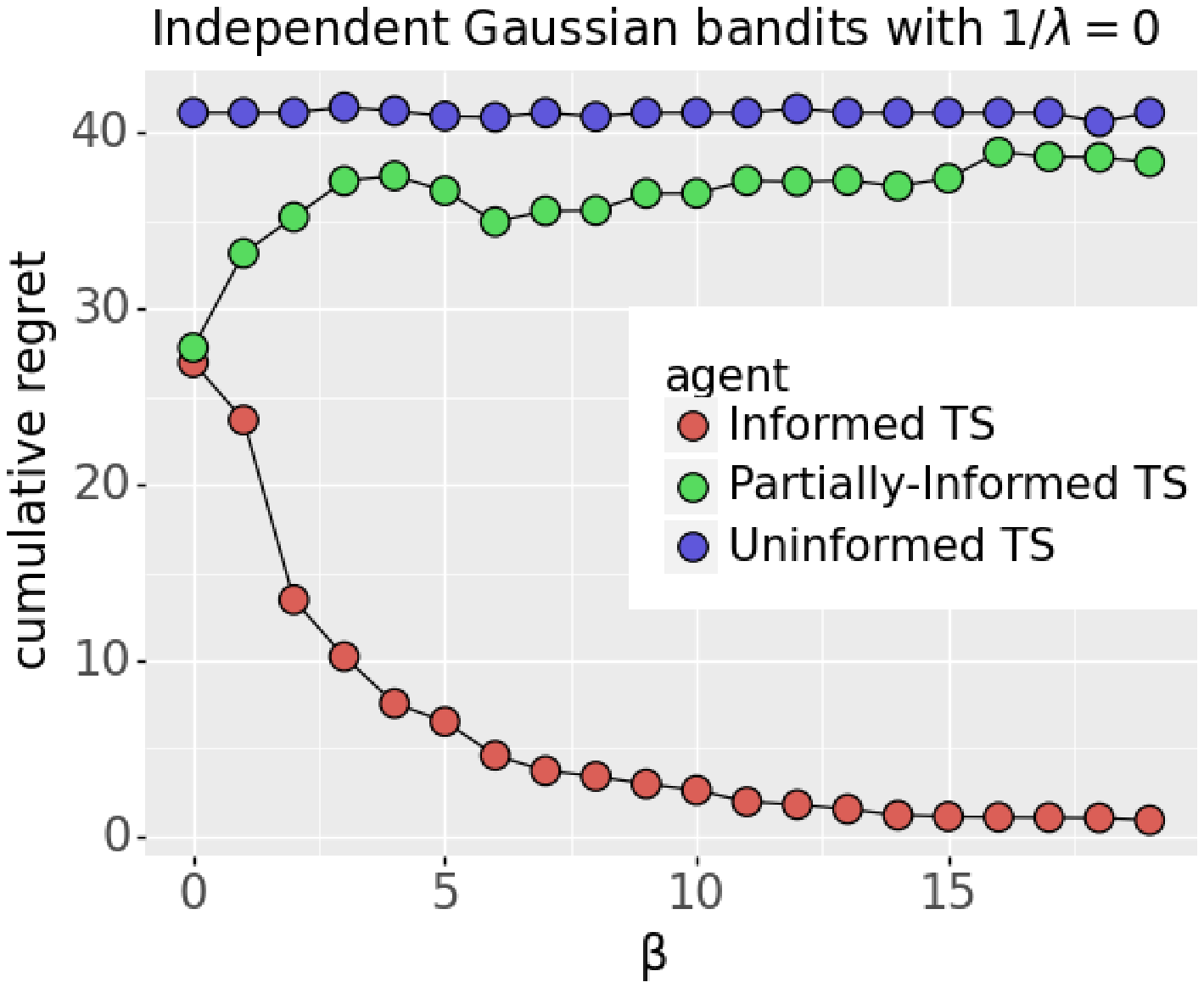}
\includegraphics[width=0.23\textwidth]{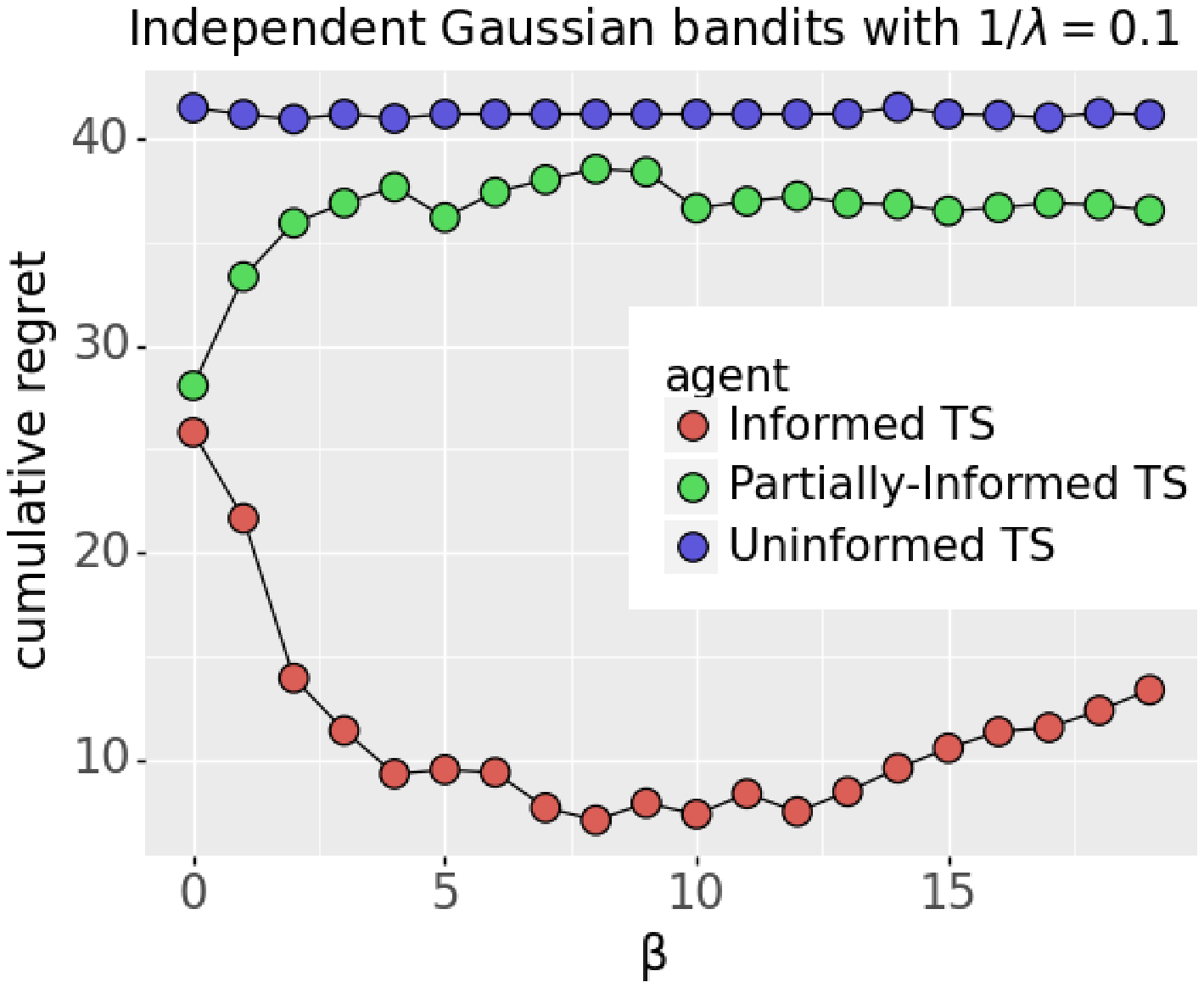}\\
\includegraphics[width=0.23\textwidth]{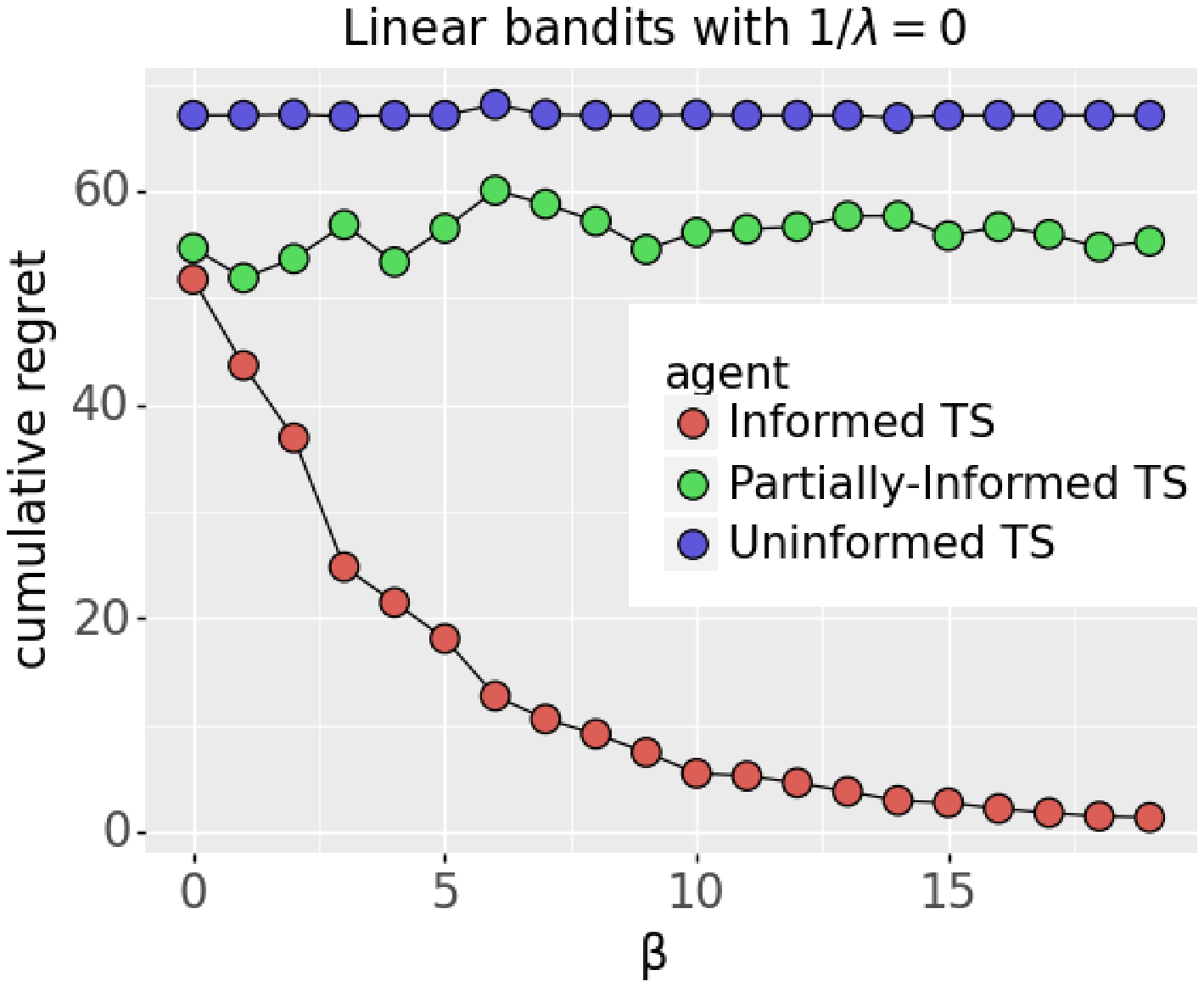}
\includegraphics[width=0.23\textwidth]{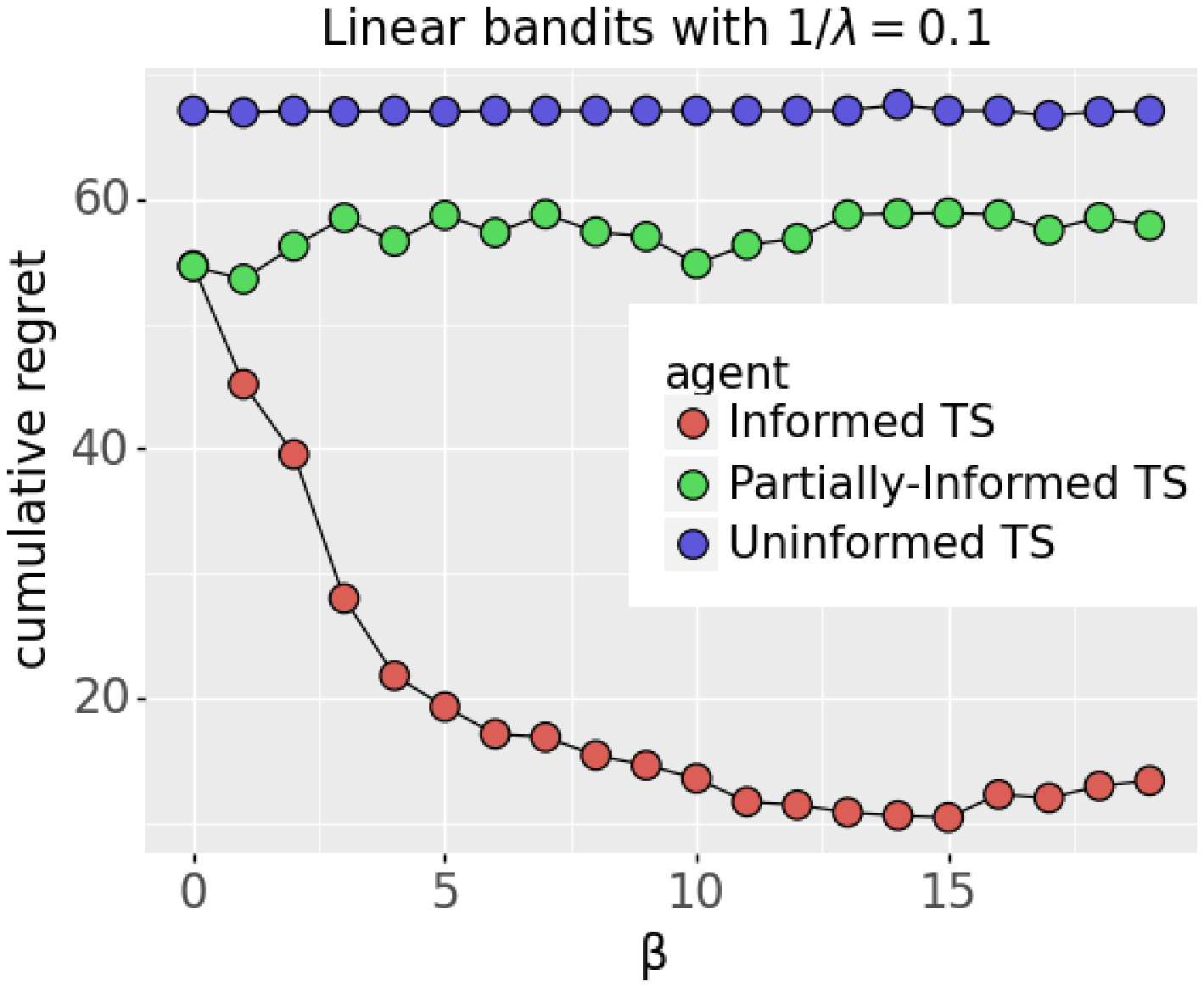}
\caption{The role of deliberateness $\beta$ with different knowledgeability $\lambda$.}  
\label{fig:fixed_lambda}
\end{figure}

\begin{figure}[h]
\includegraphics[width=0.23\textwidth]{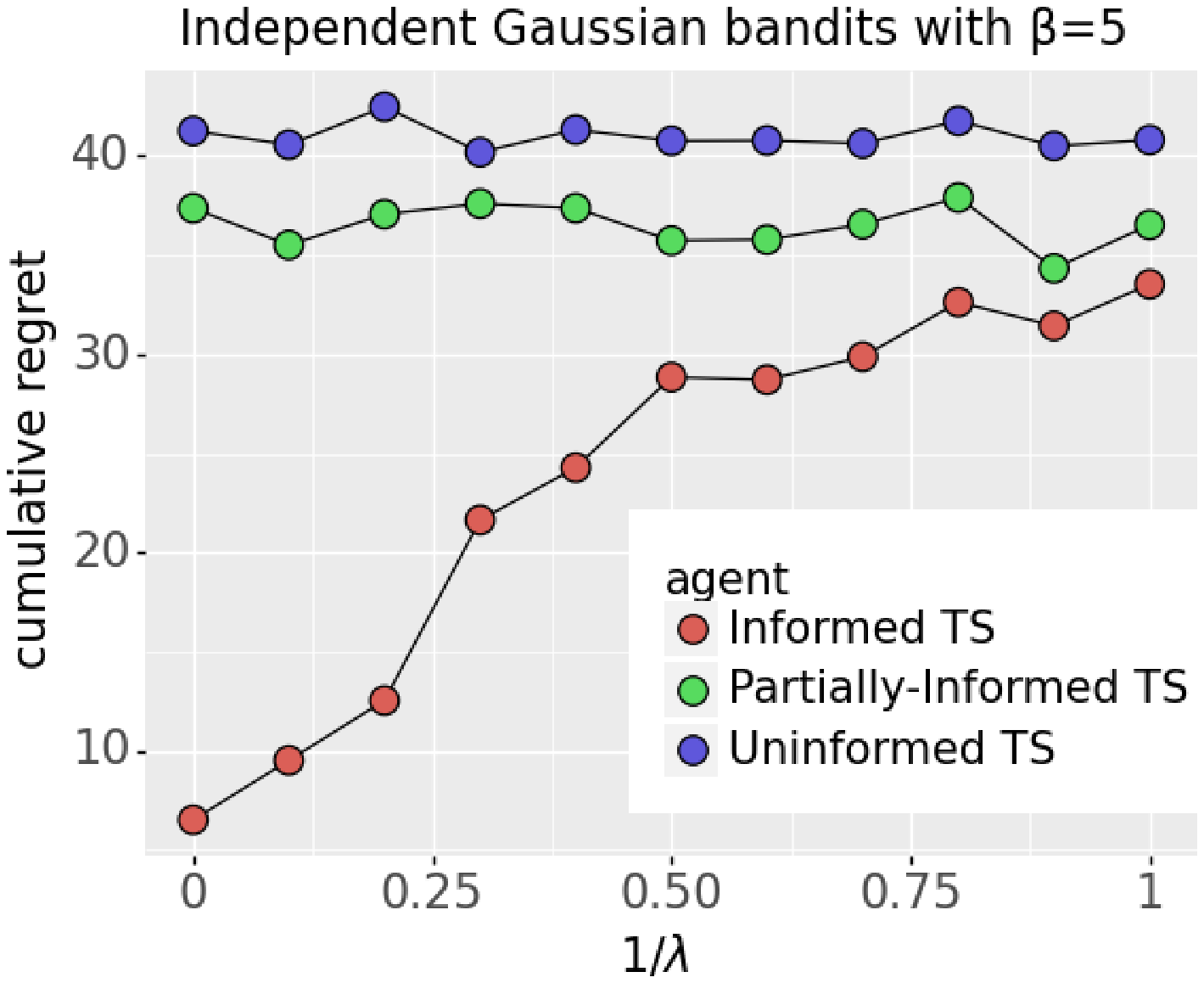}
\includegraphics[width=0.23\textwidth]{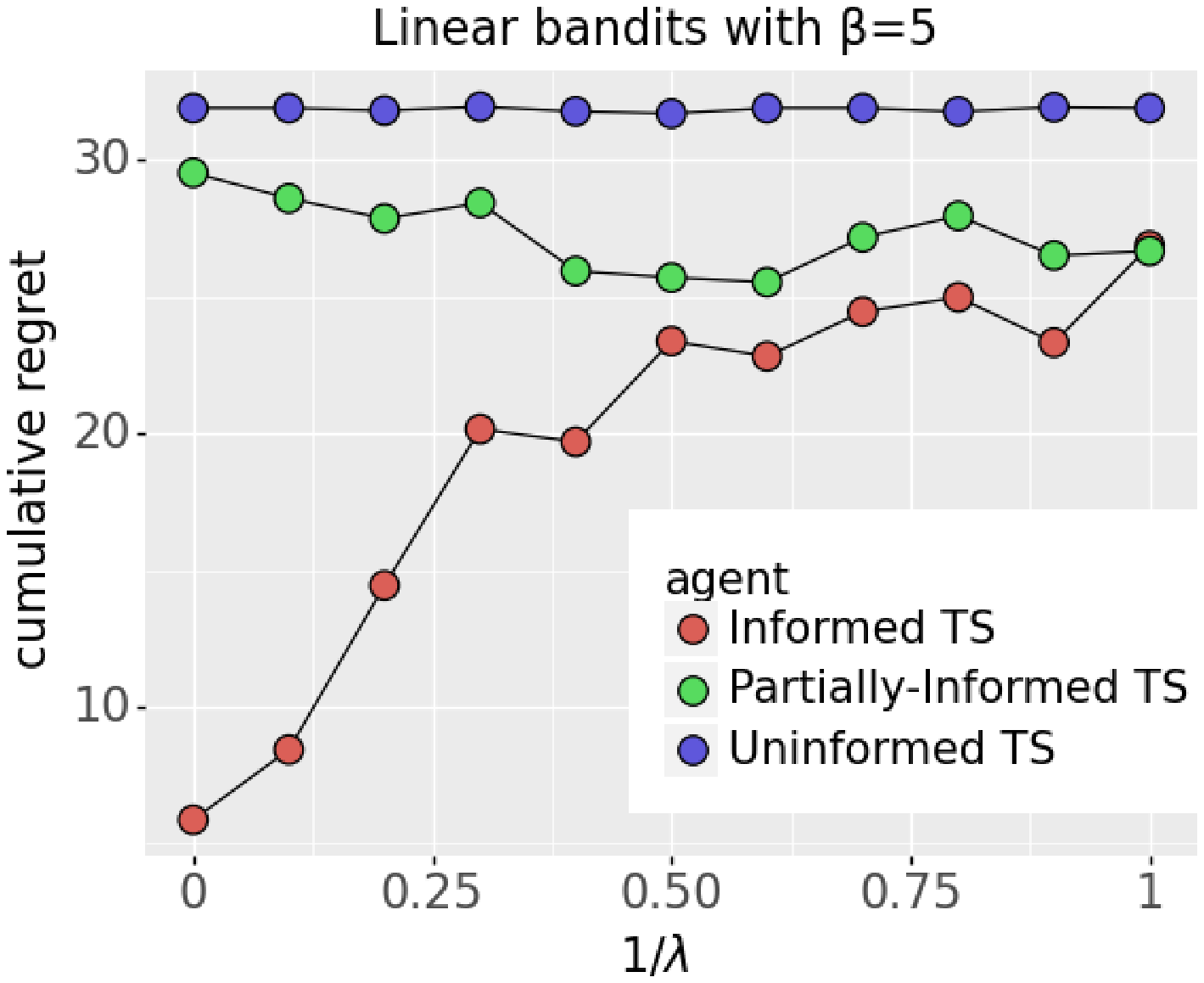}
\caption{The role of knowledgeability $\lambda$ with deliberateness $\beta=5$.}
\label{fig:fixed_beta}
\end{figure}

We first demonstrate the role of the expert's competence level (deliberatness $\beta$ and knowledability $\lambda$). Consider a Gaussian bandit with $K=5$ independent arms 
and a linear Gaussian bandit $(K=20, d=5)$ whose actions are sampled from a $d$-dimensional unit sphere. The offline demonstration datasize is fixed at  $N=10$. Consider two scenarios: (i) Fix $1/\lambda$ and vary $\beta$; and (ii) Fix $\beta$ and vary $1/\lambda$. The results are shown in Figures \ref{fig:fixed_lambda} and \ref{fig:fixed_beta}. 

In Figure \ref{fig:fixed_lambda}, for both values of $1/\lambda = 0$ and $0.1$, partially-informed TS only has a marginal regret reduction and its performance is nearly independent of the quality of the offline data. 
In contrast, informed TS enjoys a significant regret reduction that varies as $\beta$, the deliberateness level of the expert increases. In Figure \ref{fig:fixed_beta}, we see that as $\lambda$, the knowledgeability of the expert decreases, the amount of regret reduction of informed TS decreases as well. Those empirical results support our main argument that the amount of regret reduction achieved by Algorithm \ref{alg:uncertainty} by use of offline data depends on the \emph{quality} of demonstrations. 

\subsection{Robustness to Model Misspecifications}\label{sec:experiment_mis}
\begin{figure}[h]
\includegraphics[width=0.23\textwidth]{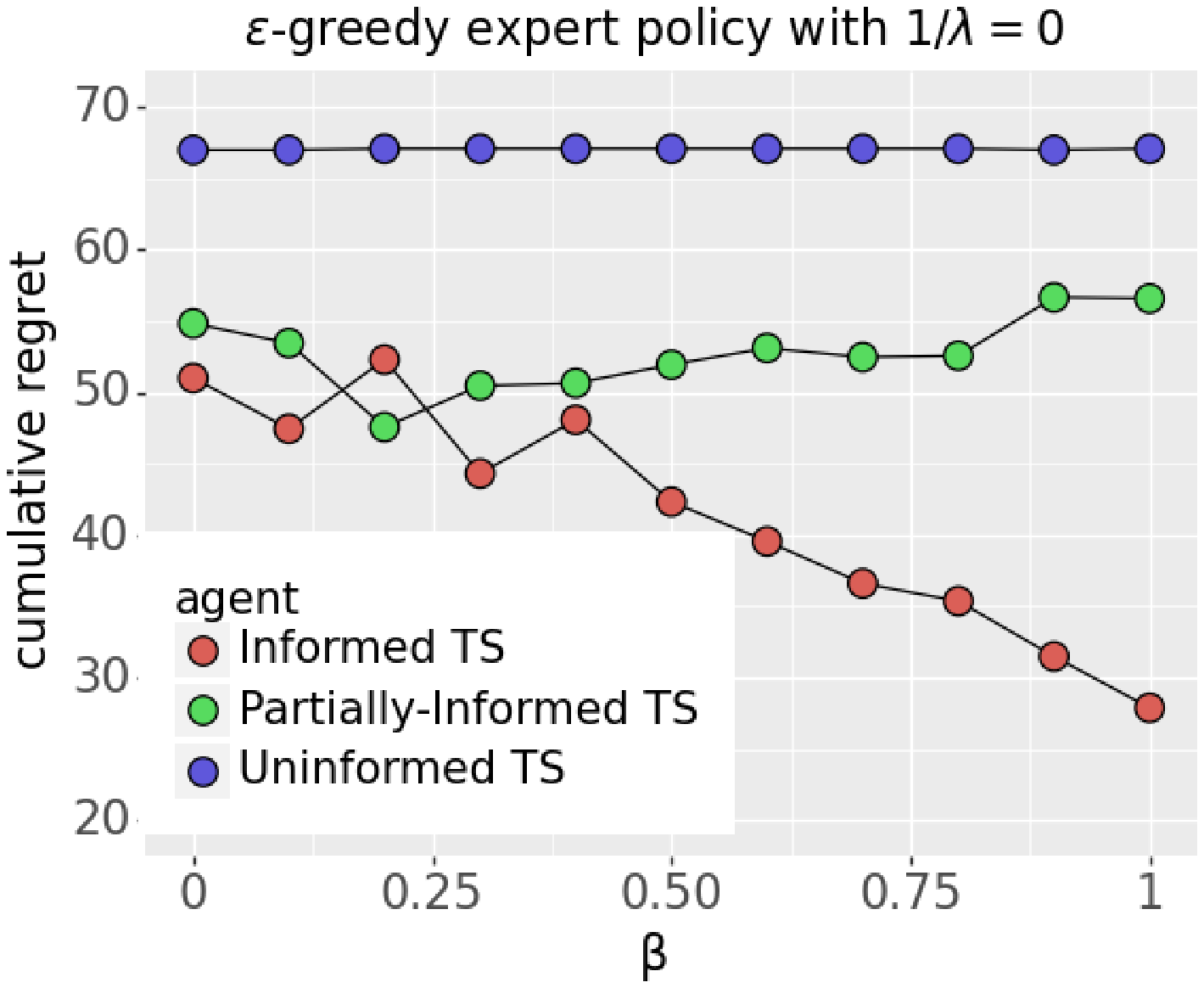}
\includegraphics[width=0.23\textwidth]{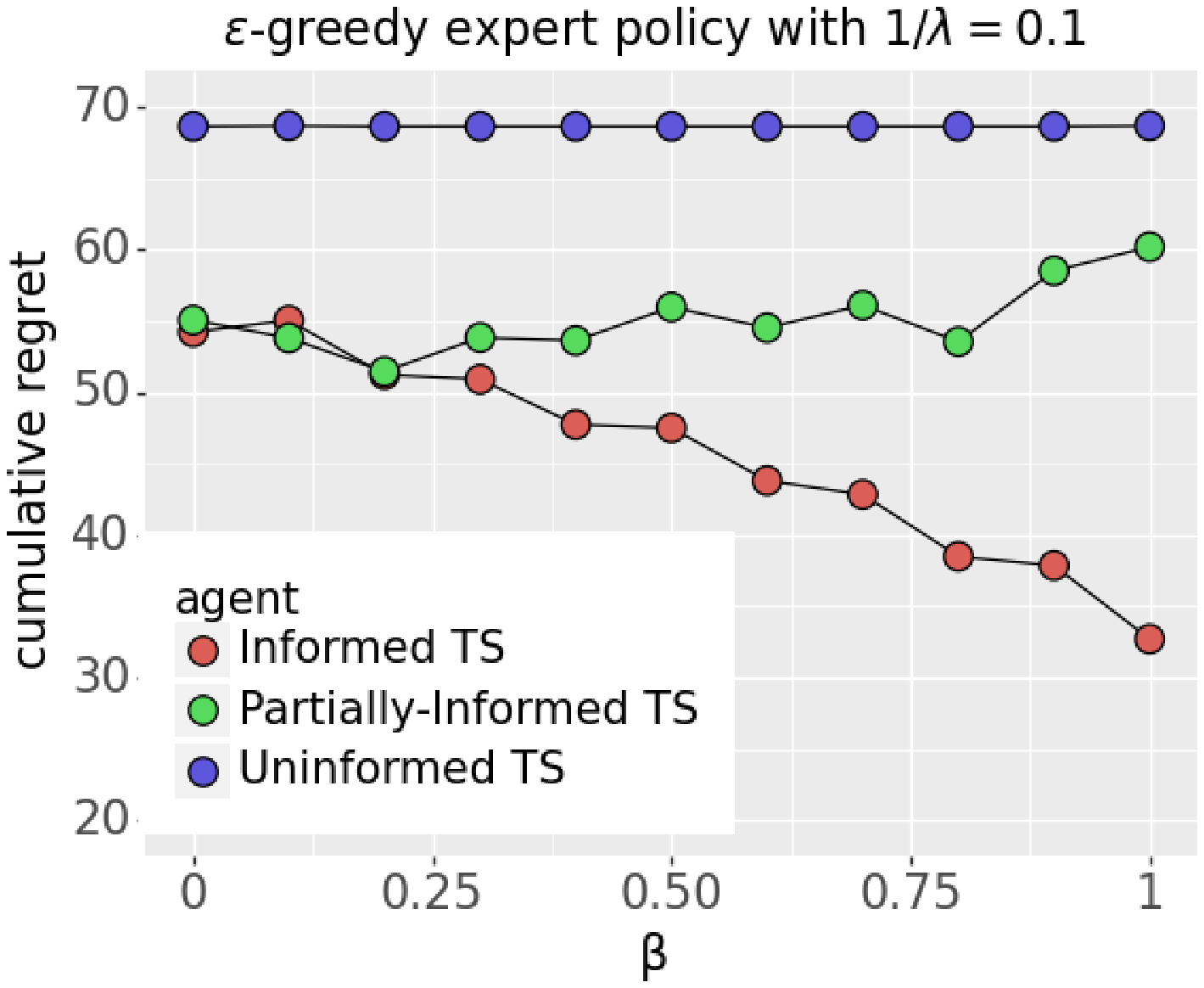}
\caption{Misspecified expert policy for linear bandits.}
\label{fig:misspecified_policy}
\end{figure}

Although the loss function in Lemma \ref{lemma:MAP} is derived from a softmax expert policy (Eq.~\eqref{eq:softmaxactions}), we would like to test if the proposed algorithm is robust to the following types of model misspecifications: 

1) \emph{Expert policy used to genetrate offline data. } 
While the algorithm assumes Eq.~\eqref{eq:softmaxactions} as its generative model, we generate the dataset by a different $\epsilon$-greedy policy introduced in Remark \ref{remark_epsilon_greedy}: for $\beta\in[0, 1]$,
        \begin{equation*}
       \phi_{\beta, \lambda}(\bar A_n=a|\vartheta) = \beta\mathbb I\left(a= \argmax_{a\in[K]}a^{\top}\vartheta\right) + (1-\beta)/K\,,
    \end{equation*}
where the expert's knowledge still takes the form of a vector $\vartheta\sim\cN(\theta, \bI_d / \lambda^2)$ conditioned on $\theta$. From Figure \ref{fig:misspecified_policy}, we can see that although the parametric form of the expert policy is misspecified, the informed TS algorithm still significantly outperforms the two baseline algorithms.

\begin{figure}[h]
\includegraphics[width=0.23\textwidth]{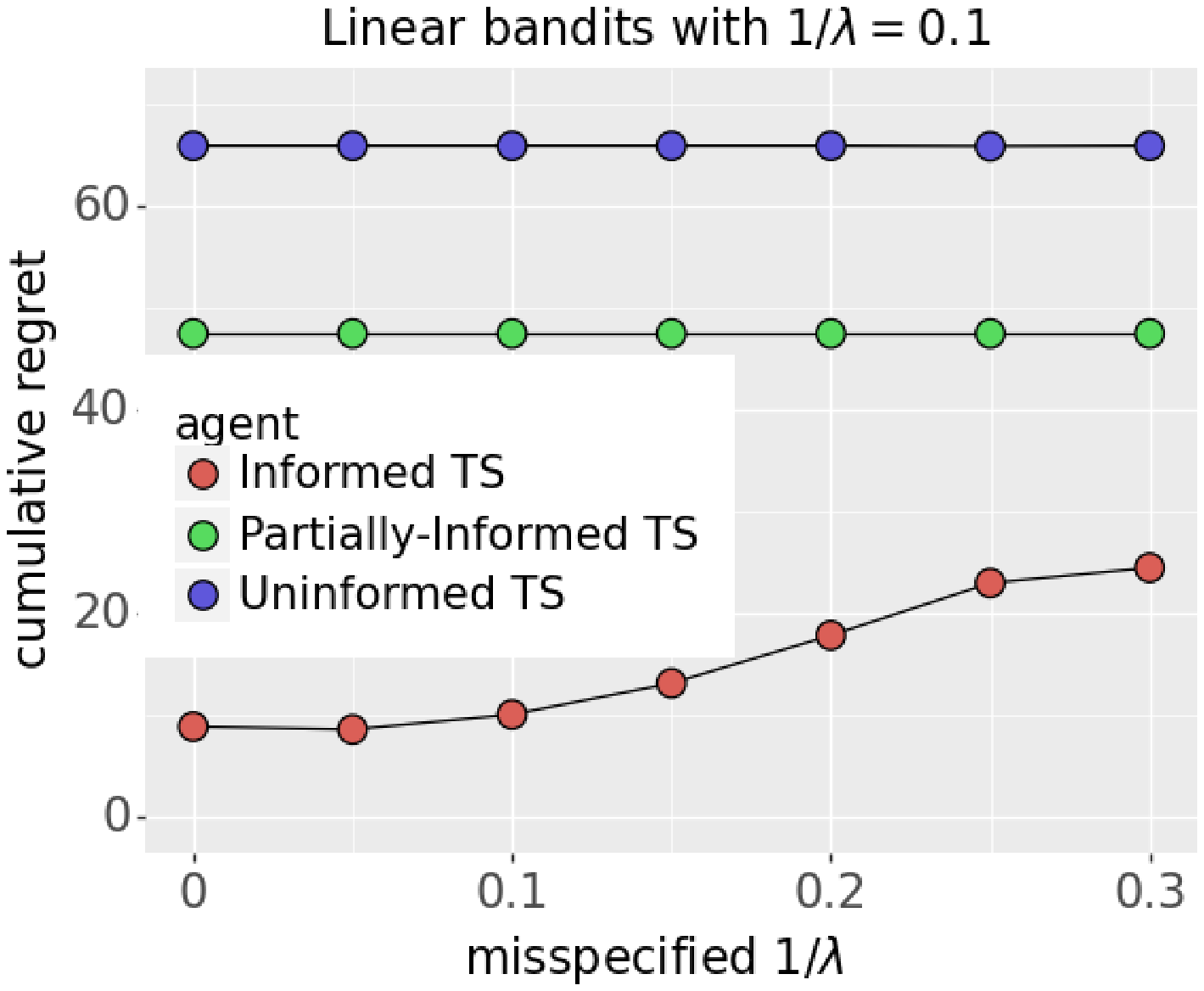}
\includegraphics[width=0.23\textwidth]{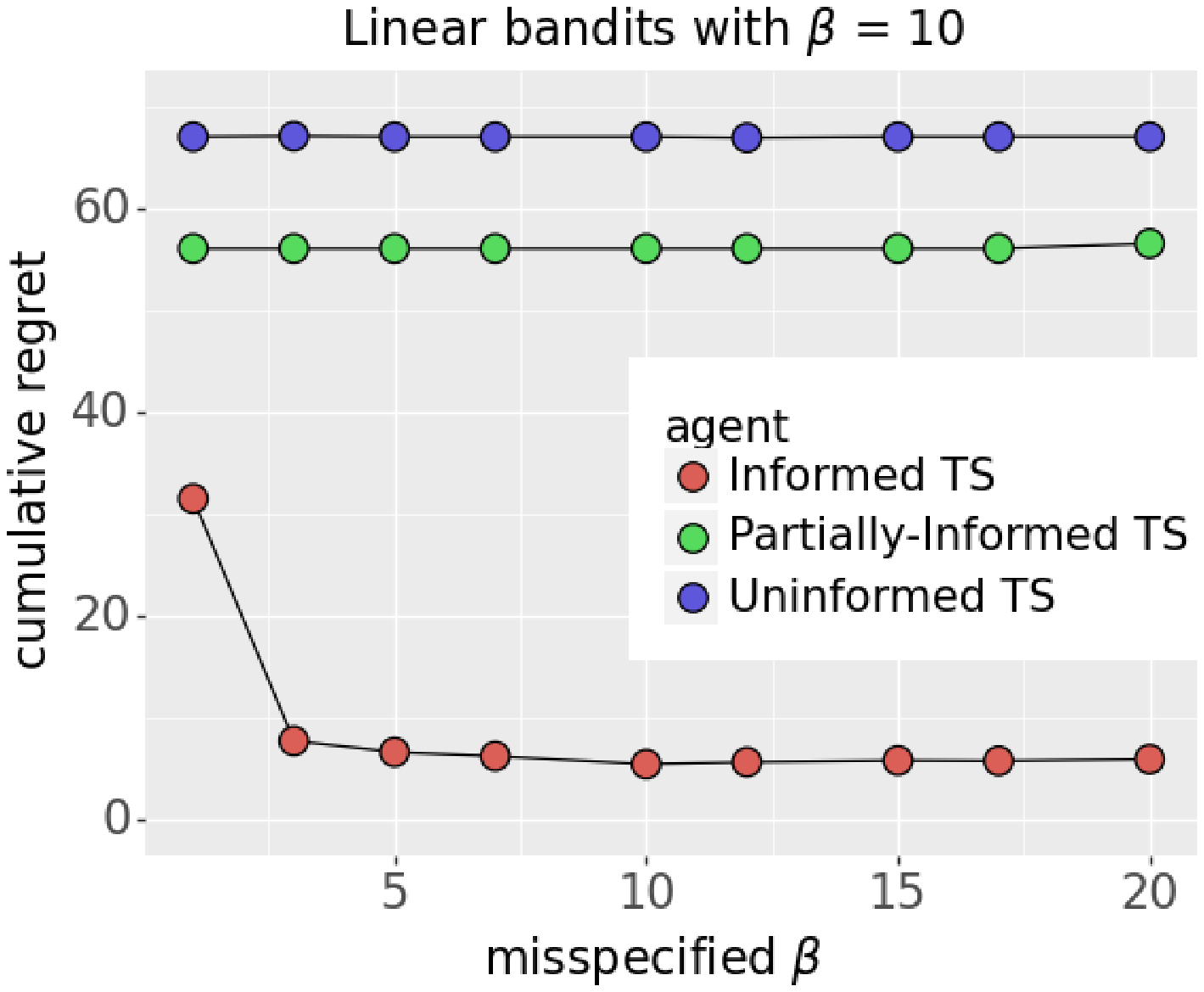}
\caption{Misspecified $\lambda$ and $\beta$.}
\label{fig:misspecified_lambda}
\end{figure}

2)  \emph{Misspecified competence level.} First, we generate the offline data with the true knowledgeability parameter $\lambda=0.1$ but the algorithm uses a misspecified $\lambda$ ranging from 0 to 1. Second, we generate the offline data with the true deliberateness parameter $\beta=10$ but the algorithm uses a misspecified $\beta$ ranging from 1 to 20. Figure \ref{fig:misspecified_lambda} shows that although the performance of informed TS decreases as the degree of misspecification increases, our algorithm still significantly outperforms the two baseline algorithms.   

\subsection{Unknown Competence Level}
\begin{figure}[h]
\centering
\includegraphics[width=0.3\textwidth]{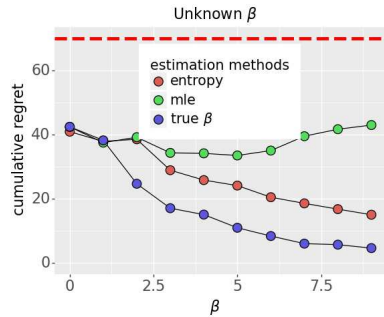}
\caption{Comparing different methods for estimating $\beta$. True $\beta$ means informed TS with known $\beta$. The red horizontal dashed line represents uninformed TS baseline.}
\label{fig:unknown_beta}
\end{figure}

We evaluate the empirical performance of the competence level estimation methods introduced in Section \ref{sec:est_competence}. We still consider a linear Gaussian bandit with $1/\lambda=0$. We compare entropy-based method and MLE based method for informed TS with two baselines: one is to plug-in true $\beta$ for informed TS; another one is the uninformed TS. As shown in Figure \ref{fig:unknown_beta}, although the performance degrades when we estimate $\beta$, our methods are still significantly outperforming the uninformed TS baseline. MLE does not perform well for a large $\beta$ since it will suffer from unbalanced offline data when computing the regularised least square estimate.

\begin{figure}
\centering
\includegraphics[width=0.3\textwidth]{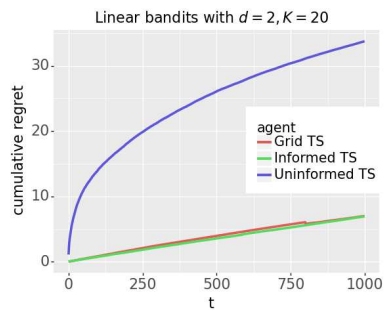}
\caption{Evaluating the approximation quality using grid TS.}
\label{fig:approx_quality}
\end{figure}
\subsection{Evaluating the Approximation Quality} 
Bayesian boostrapping is introduced because it can be computationally challenging to obtain samples from the exact posterior distribution. That immediately raises the question about how good such an approximation is. To evaluate this, we consider the $d=2$ case where we can compute a nearly exact posterior distribution by a brute-force method that relies on discretization of the parameter space. We call the algorithm as \textit{Grid-TS}. The results are shown in Figure \ref{fig:approx_quality}. We see that the regret of iTS is very close to that of Grid-TS (which works with a near exact posterior distribution). Thus the approximation quality of the Bayesian bootstrapping method is reasonably good.

%% file: conclusion.tex
\section{Conclusions and Future Work}\label{sec:conclusions}

In this paper, we have investigated how offline demonstrations can be used to improve online learning performance. It is natural to expect that use of offline dataset will result in better online learning performance. \textit{The question is how, and by how much?} Our experimental work shows that if the offline dataset is used in a naive manner, it results in only a marginal regret reduction. However, when the dataset is used by the learning agent in a more ``informed manner'', i.e., it accounts for varying levels of an expert's competence, a notion we introduce, the reduction in cumulative regret can be substantial: the higher the quality of the demonstration, the more the reduction.  

The goal of this paper is to yield insights on how to use the offline data to improve online learning. To that end, we have used a prototypical model of online learning - finite multi-armed bandits, and assumed a simple and mathematically convenient generative model for the expert policy. However, we note that the potential to generalize to more sophisticated settings and models exists. This includes linear bandits, MDPs, and more general generative models of expert policies. We will explore such directions in future work, and hope that it will inspire other researchers to explore further development of this very interesting and new research direction.

%% file: appendix.tex
\appendix
\onecolumn
\section{Proof of Lemma \ref{lemma:decomp1}}\label{sec:proof_lemma1}
We first define an information ratio with respect to a set $\cU$:
\begin{equation*}
     \Gamma_t(\cU) = \frac{\left(\sum_{a\in\cU}\mathbb P_t\left(A^*=a\right)\Delta_t(a)\right)^2}{\mathbb I_t(A^*; (A_t, R_{t}))}\,.
\end{equation*}
Applying Cauchy–Schwarz inequality, we have 
\begin{equation*}
    \begin{split}
\mathbb E\left[\sum_{t=1}^T\sum_{a\in\cU}\mathbb P_t(A^*=a)\Delta_t(a)\right] &=\mathbb E\left[\sum_{t=1}^T\frac{\sum_{a\in\cU}\mathbb P_t(A^*=a)\Delta_t(a)}{\sqrt{\mathbb I_t(A^*; (A_t, R_{t}))}}\sqrt{\mathbb I_t(A^*; (A_t, R_{t}))}\right]\\
&\leq \sqrt{\mathbb E\left[\sum_{t=1}^T\frac{\left(\sum_{a\in\cU}\mathbb P_t(A^*=a)\Delta_t(a)\right)^2}{\mathbb I_t(A^*; (A_t, R_{t}))}\right]\sum_{t=1}^T\mathbb E\left[\mathbb I_t(A^*; (A_t, R_{t}))\right]}\,.
    \end{split}
\end{equation*}
Using the chain rule of mutual information, 
\begin{equation*}
    \begin{split}
        \sum_{t=1}^T\mathbb E\left[\mathbb I_t(A^*; (A_t, R_{t}))\right] = \mathbb I(A^*; \cD_T)\leq \cH(A^*)\,.
    \end{split}
\end{equation*}
Then we have 
\begin{equation}\label{eqn:bound42}
    \mathbb E\left[\sum_{t=1}^T\sum_{a\in\cU}\mathbb P_t(A^*=a)\Delta_t(a)\right] \leq \sqrt{\sum_{t=1}^T\mathbb E[\Gamma_t(\cU)]\cH(A^*)}\,.
\end{equation}

For the information ratio $\mathbb E[\Gamma_t(\cU)]$, applying Cauchy–Schwarz inequality over the set $\cU$ and following the step in Eq.~\eqref{eqn:standard_bound},
\begin{equation*}
    \begin{split}
       \left(\sum_{a\in\cU}\mathbb P_t(A^*=a)\Delta_t(a)\right)^2
  \leq |\cU|\frac{\mathbb I_t(A^*; (A_t, R_{t}))}{2}\,,
    \end{split}
\end{equation*}
which implies $\mathbb E[\Gamma_t(\cU)]\leq \mathbb E[|\cU|]$ for any $t\in[T]$. For the entropy part $\cH(A^*)$, by the definition of Shannon entropy,
\begin{equation}\label{eqn:entropy1}
    \begin{split}
      \cH\left(A^*\right) &= \mathbb E\left[\cH(\mathbb P(A^*\in\cdot|\cU))\right]\\
      &= \mathbb E\left[-\sum_{a\in\cU}\mathbb P(A^*=a)\log(\mathbb P(A^*=a))-\sum_{a\notin\cU}\mathbb P(A^*=a)\log(\mathbb P(A^*=a))\right]
      \,.
    \end{split}
\end{equation}
\begin{itemize}
    \item The first term can be bounded by the uniform bound on the entropy of a probability distribution:
\begin{equation}\label{eqn:entropy2}
\begin{split}
\mathbb E\left[ -\sum_{a\in\cU}\mathbb P(A^*=a)\log(\mathbb P(A^*=a))\right]\leq \mathbb E\left[\log\left(|\cU|\right)\right]\leq \log\left(\mathbb E[|\cU|]\right)\,.
   \end{split}
\end{equation}
where the second inequality uses Jenson's inequality. 
\item For the second term, we use the following fact and the definition of $\cU$ in Eq.~\eqref{def:informative_set}:
\begin{equation}\label{eqn:entropy3}
   \begin{split}
 &\mathbb E\left[ -\sum_{a\notin\cU}\mathbb P(A^*=a)\log(\mathbb P(A^*=a)) \right] \\ 
 = &\mathbb E\left[ \varepsilon \sum_{a\notin\cU}\frac{\mathbb P(A^*=a)}{\varepsilon}\log\left(\frac{\varepsilon}{\mathbb P(A^*=a)}\right)\right] - \mathbb E\left[\sum_{a\notin\cU}\mathbb P(A^*=a)\log(\varepsilon)\right] \\
  \leq& \mathbb E\left[\varepsilon \log\left(K-|\cU|\right)\right]+\mathbb E\left[\sum_{a\notin\cU}\mathbb P(A^*=a)\right]\log(1/\varepsilon)\\
  \leq&  \varepsilon \log(K)+\varepsilon\log(1/\varepsilon) = \varepsilon \log (K/\varepsilon)
\,.    \end{split}
\end{equation}
\end{itemize}
Putting Eqs.~\eqref{eqn:entropy1}-\eqref{eqn:entropy3} together,
\begin{equation*}
   \cH(A^*)\leq \log\left(\mathbb E[|\cU|]\right) +\varepsilon\log\left(K/\varepsilon\right)\,. 
\end{equation*}
Plugging the above bounds into Eq.~\eqref{eqn:bound42},
\begin{equation*}
  \mathbb E\left[\sum_{t=1}^T\sum_{a\in\cU}\mathbb P_t(A^*=a)\Delta_t(a)\right] \leq \sqrt{T\mathbb E[|\cU|]\left(\log\left(\mathbb E[|\cU|]\right)+\varepsilon\log\left(K/\varepsilon\right)\right)}\,.
\end{equation*}
We now reach the conclusion.

\section{Proof of Lemma \ref{lemma:decomp2}}\label{sec:proof_lemma2}
By the definition of $\Delta_t(a)$,
\begin{equation*}
    \begin{split}
          \mathbb E\left[\sum_{t=1}^T\sum_{a\notin\cU}\mathbb P_t\left(A^*=a\right)\Delta_t(a)\right] =   \mathbb E\left[\sum_{t=1}^T\sum_{a\notin\cU}\mathbb P_t\left(A^*=a\right)\left(\mathbb E_t[\langle a, \theta\rangle|A^*=a]-\mathbb E_t[\langle a, \theta\rangle]\right)\right]\,.
    \end{split}
\end{equation*}
First, we note that 
\begin{equation*}
    \begin{split}
        \sum_{a\notin\cU}\mathbb P_t(A^*=a)\mathbb E_t\left[\langle a,\theta\rangle|A^*=a\right]&\leq  \sum_{a\in\cA}\mathbb P_t(A^*=a)\mathbb E_t\left[\langle a,\theta\rangle| A^*=a\right]\\
        &=\mathbb E_t\left[\langle A^*, \theta\rangle\right] = \mathbb E_t\left[\max_{a}\langle a, \theta\rangle\right]\,.
    \end{split}
\end{equation*}
Therefore,
\begin{equation*}
     \mathbb E\left[\sum_{t=1}^T\sum_{a\notin\cU}\mathbb P_t\left(A^*=a\right)\Delta_t(a)\right] \leq \sum_{t=1}^T \mathbb E\left[\sum_{a\notin\cU}\mathbb P_t(A^*=a)\mathbb E_t\left[\max_a\langle a,\theta\rangle-\min_a\langle a, \theta\rangle\right]\right]\,.
\end{equation*}
By Cauchy-Schwarz inequality, 
\begin{equation*}
    \begin{split}
    \mathbb E\left[\sum_{a\notin\cU}\mathbb P_t(A^*=a)\mathbb E_t\left[\max_a\langle a,\theta\rangle-\min_a\langle a, \theta\rangle\right]\right]&\leq \sqrt{\mathbb E\left[\left(\sum_{a\notin\cU}\mathbb P_t(A^*=a)\right)^2\right]\mathbb E\left[\left(\mathbb E_t\left[\max_a\langle a,\theta\rangle-\min_a\langle a, \theta\rangle\right]\right)^2\right]}\\
    &\leq \sqrt{\left(\mathbb E\left[\sum_{a\notin\cU}\mathbb P_t(A^*=a)\right]\right)^2\left(\mathbb E\left[\mathbb E_t\left[\max_a\langle a,\theta\rangle-\min_a\langle a, \theta\rangle\right]\right]\right)^2} \\
    &= \mathbb E\left[\sum_{a\notin\cU}\mathbb P_t(A^*=a)\right]\mathbb E\left[\mathbb E_t\left[\max_a\langle a,\theta\rangle-\min_a\langle a, \theta\rangle\right]\right]
    \end{split}
\end{equation*}
where we use Jenson's inequality for the second inequality. According to the tower property of conditional expectation, we have 
\begin{equation*}
    \begin{split}
    \mathbb E\left[\mathbb E_t\left[\max_a\langle a,\theta\rangle-\min_a\langle a, \theta\rangle\right]\right] =  \mathbb E\left[\max_a\langle a,\theta\rangle-\min_a\langle a, \theta\rangle\right]\,,
    \end{split}
\end{equation*}
and
\begin{equation*}
    \begin{split}
   \mathbb E\left[\sum_{a\notin\cU}\mathbb P_t(A^*=a)\right] &= \mathbb E\left[\sum_{a\notin\cU}\mathbb E\left[\mathbb I\left(A^*=a\right)|\cD_t \right]\right]  \\
     &=\mathbb E\left[\sum_{a\notin\cU}\mathbb I\left(A^*=a\right)\right] = \mathbb P(A^*\notin \cU)\,.
    \end{split}
\end{equation*}
When $\mathbb E\left[\max_a\langle a,\theta\rangle-\min_a\langle a, \theta\rangle\right]$ is bounded by $C_1$, we reach the conclusion.

\section{Proof of Lemma \ref{lemma:lemma1}}\label{sec:proof_lemma1}

We split the proof of Lemma \ref{lemma:lemma1} into two parts: prove $\cU_{A}$ is $(1-f_1)$-informative and prove $
\mathbb E[|\cU_{A}|]\leq f_2
$.
\subsection{Prove $\cU_{A}$ is $(1-f_1)$-informative}\label{sec:proof_lemma1_part1}
Based on the assumption of the offline data generating process, we know that conditioned on $\vartheta$, $\bar A_n$ is independent of $\bar A_{n'}$ for any $n\neq n'$. This implies
\begin{equation}\label{eqn1}
    \begin{split}
        \mathbb P\left(A^*\notin\cU_{A}\right)& = \mathbb P\left(\bar A_n\neq A^*, \forall n\in[N]\right) = \mathbb P\left(\bigcap_{n=1}^N\left\{\bar A_n\neq A^*\right\}\right) \\
        &=\mathbb E\left[\mathbb P\left(\bigcap_{n=1}^N\left\{\bar A_n\neq A^*\right\}\Big|\theta, \vartheta\right)\right]
        = \mathbb E\left[\prod_{n=1}^N\mathbb P(\bar A_n\neq A^*\big|\theta, \vartheta)\right]\\
        & = \mathbb E\left[ \prod_{n=1}^N\left(1-\mathbb P\left(\bar A_n=A^*\Big|\theta, \vartheta\right)\right)\right] \,,
    \end{split}
\end{equation}
where $A^*$ is a function of $\theta$ and thus a random variable as well. According to the definition of the softmax expert policy in Eq.~\eqref{eq:softmaxactions},
\begin{equation*}
    \begin{split}
        \mathbb P(\bar A_n=A^*|\theta, \vartheta) &= \frac{\exp(\beta\vartheta^\top A^*)}{\sum_{b\in\cA}\exp(\beta \vartheta^\top b)}=\frac{1}{\sum_{b\in\cA} \exp(-\beta\langle A^*-b, \vartheta\rangle)}\\
        &=\left(\sum_{b\in\cA}\exp\left(\beta \langle A^*-b, \theta-\vartheta\rangle-\beta \langle A^*-b, \theta\rangle\right)\right)^{-1}\\
        &\geq\left(\sum_{b\in\cA}\exp\left(\beta \|A^*-b\|_1\| \vartheta-\theta\|_{\infty}-\beta \langle A^*-b, \theta\rangle\right)\right)^{-1}\,.
    \end{split}
\end{equation*}
For multi-armed bandits, $\| A^*-b\|_1\leq 1$ almost surely for any $b\in\cA$ so 
\begin{equation*}
     \mathbb P(\bar A_n=A^*|\theta, \vartheta) \geq \left(\sum_{b\in\cA}\exp\left(\beta \| \vartheta-\theta\|_{\infty}-\beta \langle A^*-b, \theta\rangle\right)\right)^{-1}\,.
\end{equation*}

Since $\vartheta-\theta\sim N(0, \bI_K/\lambda^2)$, using standard Hoeffding's bound \cite{vershynin2010introduction} implies 
\begin{equation*}
    \mathbb P\left(\|\vartheta-\theta\|_{\infty}\geq t\right)\leq K\exp\left(-\frac{t^2\lambda^2}{2}\right)\,.
\end{equation*}
Set $t=\sqrt{2\log(TK)}/\lambda$ and define an event $\cE_1:=\{\|\vartheta-\theta\|_{\infty}\leq \sqrt{2\log(TK)}/\lambda\}$ such that $\mathbb P(\cE_1^c)\leq 1/T$. We decompose Eq.~\eqref{eqn1} according to $\cE_1$:
\begin{equation}\label{eqn2}
    \begin{split}
         \mathbb P\left(A^*\notin\cU_{A}\right)&\leq \mathbb E\left[ \prod_{n=1}^N\left(1-\mathbb P\left(\bar A_n=A^*\Big|\theta, \vartheta\right)\right)\mathbb I(\cE_1)\right] + \mathbb P(\cE_1^c)  \\
         &\leq \mathbb E\left[\prod_{n=1}^N\left(1-\left(\exp\left(\frac{\beta \sqrt{2\log(TK)}}{\lambda}\right)\sum_{b\in\cA}\exp\left(-\beta \langle A^*-b, \theta\rangle\right)\right)^{-1}\right)\right]+ \frac{1}{T}\,.
    \end{split}
\end{equation} 

Let us define a set
\begin{equation}\label{def:set_B}
    \cB=\left\{a:\langle A^*-a, \theta\rangle\leq \Delta\right\}\,,
\end{equation}
where $\Delta$ will be chosen later. Then we can further decompose Eq.~\eqref{eqn2} as 
\begin{equation*}
\begin{split}
   & \mathbb P\left(A^*\notin\cU_{A}\right)\\
   &\leq \mathbb E\left[\prod_{n=1}^N\left(1-\left(\exp\left(\frac{\beta \sqrt{2\log(TK)}}{\lambda}\right)\left(\sum_{b\in\cB}\exp\left(-\beta \langle A^*-b, \theta\rangle\right)+\sum_{b\notin\cB}\exp\left(-\beta \langle A^*-b, \theta\rangle\right)\right)\right)^{-1}\right)\right]+ \frac{1}{T}\,.
    \end{split}
\end{equation*}

For $b\notin \cB$, we have $\exp(-\beta \langle A^*-b, \theta\rangle)\leq \exp(-\beta\Delta)$. For $b\in\cB$, we have $\exp(-\beta  \langle A^*-b, \theta\rangle)\leq \exp(-\beta 0)=1.$
Putting the above together, we can have 
\begin{equation}\label{eqn3}
    \sum_{b\in \cB}\exp(-\beta\langle A^*-b, \theta\rangle)+\sum_{b\notin \cB}\exp(-\beta\langle A^*-b, \theta\rangle)\leq |\cB|+(K-|\cB|)\exp(-\beta\Delta)\leq |\cB|+K\exp(-\beta\Delta)\,,
\end{equation}
where $|\cB|=\sum_{b}\mathbb I(\langle A^*-b, \theta\rangle\leq \Delta)$ is a random variable.
Putting Eqs.~\eqref{eqn2}-\eqref{eqn3} together implies
\begin{equation}\label{eqn:bound1}
\begin{split}
     \mathbb P\left(A^*\notin\cU_{A}\right)&\leq  \mathbb E\left[ \left(1-\frac{\exp\left(-\beta \sqrt{2\log(TK)}/\lambda\right)}{|\cB|+K\exp(-\beta\Delta)}\right)^N\right]+\frac{1}{T}\\
     &=\sum_{k=0}^{K-1}\mathbb P\left(|\cB|-1=k\right)\left(1-\frac{\exp\left(-\beta \sqrt{2\log(TK)}/\lambda\right)}{k+K\exp(-\beta\Delta)}\right)^N+\frac{1}{T}\\
     &=\sum_{k=0}^{K-1}\sum_{a\in\cA}\mathbb P\left(|\cB|-1=k|A^*=a\right)\mathbb P\left(A^*=a\right)\left(1-\frac{\exp\left(-\beta \sqrt{2\log(TK)}/\lambda\right)}{k+K\exp(-\beta\Delta)}\right)^N+\frac{1}{T}\,,
\end{split}
\end{equation}
where we use the change-of-variables formula for the push-forward measure for the first equation. 

Next we study the distribution of $|\cB|-1$ conditional on $A^*$. Without loss of generality, we first consider the conditional distribution with conditioning on $A^*=a_1$:
\begin{equation}\label{eqn:B_dis}
    \begin{split}
        \mathbb P\left(|\cB|-1=k|A^*=a_1\right) &= \mathbb P\left(\sum_{a\in\cA}\mathbb I\left(\langle A^*-a, \theta\rangle\leq \Delta\right)-1=k\Big|A^*=a_1\right)\\
        &=\frac{1}{\mathbb P\left(A^*=a_1\right)}\mathbb P\left(\sum_{a\in\cA}\mathbb I\left(\langle A^*-a, \theta\rangle\leq \Delta\right)-1=k, A^*=a_1\right)\\
        &=\frac{1}{\mathbb P\left(A^*=a_1\right)}\mathbb P\left(\sum_{a\in\cA}\mathbb I\left(\theta_a\geq \theta_1-\Delta\right)-1=k, \bigcap_{a\in\cA}\{\theta_1\geq \theta_a\}\right)\\
        &=\frac{1}{\mathbb P\left(A^*=a_1\right)}\int_R \binom{K-1}{k}\left[\int_{\theta_1-\Delta}^{\theta_1}\d \rho(\theta)\right]^k\left[\int_{-\infty}^{\theta_1-\Delta}\d\rho(\theta)\right]^{K-1-k} \d \rho(\theta_1)\,,
    \end{split}
\end{equation}
where $\rho(\cdot)$ is the univariate Gaussian distribution and $\theta_a = a^{\top}\theta$ for any $a\in\cA$.
By defining $$F(\theta_1) =\int_{-\infty}^{\theta_1}(2\pi)^{-1/2}\exp(-x^2/2)\d x\,,$$ we have 
\begin{equation*}
    \int_{\theta_1-\Delta}^{\theta_1}\frac{\d \rho(\theta)}{F(\theta_1)} + \int^{\theta_1-\Delta}_{\infty}\frac{\d \rho(\theta)}{F(\theta_1)}=1\,.
\end{equation*}
We further define
\begin{equation}\label{def:q_theta}
    q(\theta_1) = \int_{\theta_1-\Delta}^{\theta_1}\frac{\d \rho(\theta)}{F(\theta_1)}\,.
\end{equation}
For a fixed $\theta_1$, let $X_{\theta_1}\sim\text{Binomial}(K-1, q(\theta_1))$.
Together with Eq.~\eqref{eqn:B_dis},
\begin{equation}\label{eqn:B_prob}
\begin{split}
     \mathbb P\left(|\cB|-1=k|A^*=a_1\right) &= \int_R \mathbb P(X_{\theta_1}=k) \frac{F(\theta_1)^{K-1}}{\mathbb P(A^*=a_1)}\d \rho(\theta_1)=\int_R \mathbb P(X_{\theta_1}=k)\d \mu(\theta_1)\,,
\end{split}
\end{equation}
where we denote
\begin{equation*}
    \d \mu(\theta_1) = \frac{F(\theta_1)^{K-1}}{\mathbb P(A^*=a_1)}\d \rho(\theta_1)\,.
\end{equation*}
Thus, $|\cB|-1$ follows a mixture of binomial distribution.
Plugging Eq.~\eqref{eqn:B_prob} into Eq.~\eqref{eqn:bound1},
\begin{equation*}
    \begin{split}
        \mathbb P\left(A^*\notin\cU_{A}\right)&\leq  \sum_{k=0}^{K-1}\sum_{a\in\cA}\int_R \mathbb P(X_{\theta_{a}}=k)\d \mu(\theta_a)\mathbb P\left(A^*=a\right)\left(1-\frac{\exp\left(-\beta \sqrt{2\log(TK)}/\lambda\right)}{k+K\exp(-\beta\Delta)}\right)^N+\frac{1}{T}\\
        &= \sum_{a\in\cA}\int_R\sum_{k=0}^{K-1} \mathbb P(X_{\theta_{a}}=k)\left(1-\frac{\exp\left(-\beta \sqrt{2\log(TK)}/\lambda\right)}{k+K\exp(-\beta\Delta)}\right)^N\d \mu(\theta_a)\mathbb P\left(A^*=a\right)+\frac{1}{T}\\
        &=\sum_{a\in\cA}\int_R\mathbb E\left[\left(1-\frac{\exp\left(-\beta \sqrt{2\log(TK)}/\lambda\right)}{1+X_{\theta_a}+K\exp(-\beta\Delta)}\right)^N\Big|\theta_a\right]\d \mu(\theta_a)\mathbb P\left(A^*=a\right)+\frac{1}{T}\\
        &=\int_R\mathbb E\left[\left(1-\frac{\exp\left(-\beta \sqrt{2\log(TK)}/\lambda\right)}{1+X_{\theta_1}+K\exp(-\beta\Delta)}\right)^N\Big|\theta_1\right]\d \mu(\theta_1)+\frac{1}{T}\,,
    \end{split}
\end{equation*}
where the last equation is due to $\mathbb P(A^*=a)=1/K$ for any $a\in\cA$.

For fixed $\theta_1$, 
using the Stirling's approximation for binomial tail bound,
\begin{equation*}
\begin{split}
      \mathbb P\left(X_{\theta_1}\geq (K-1)q(\theta_1)+\frac{\log(T)}{\log(1/((K-1)q(\theta_1)))}\Big |\theta_1\right)\leq \frac{1}{T}\,.
\end{split}
\end{equation*}
Define an event 
$$
\cE_2=\left\{X_{\theta_1}\leq (K-1)q(\theta_1)+\frac{\log(T)}{\log(1/((K-1)q(\theta_1)))}\right\}\,,
$$ such that 
$
    \mathbb P(\cE_2^c|\theta_1)\leq 1/T.
$
We decompose the above based on $\cE_2$:
\begin{equation}\label{eqn:A_*}
    \begin{split}
    \mathbb P\left(A^*\notin\cU_{A}\right)&\leq
  \int_R\mathbb E\left[\left(1-\frac{\exp\left(-\beta \sqrt{2\log(TK)}/\lambda\right)}{1+X_{\theta_1}+K\exp(-\beta\Delta)}\right)^N\mathbb I(\cE_1) \Big|\theta_1\right]\d \mu(\theta_1)
    +\frac{2}{T} \\
    &\leq
  \int_R\mathbb E\left[\left(1-\frac{\exp\left(-\beta \sqrt{2\log(TK)}/\lambda\right)}{1+(K-1)q(\theta_1)+\frac{\log(T)}{\log(1/((K-1)q(\theta_1)))}+K\exp(-\beta\Delta)}\right)^N\mathbb I(\cE_1) \Big|\theta_1\right]\d \mu(\theta_1)
    +\frac{2}{T}\,,
     \end{split}
\end{equation}
where the expectation is with respect to $\theta_1$ under measure $\mu(\cdot)$.
It remains to derive an upper bound $q(\theta_1)$ which is defined in Eq.~\eqref{def:q_theta}. 
\begin{itemize}
    \item First, we have 
\begin{equation*}
    \begin{split}
        \mathbb P\left(\theta_1\leq x|A^*=a_1\right)&= \frac{ \mathbb P\left(\theta_1\leq x, A^*=a_1\right)}{\mathbb P\left(A^*=a_1\right)}\\
        &=\frac{ \mathbb P\left(\theta_1\leq x, \bigcap_{a\neq a_1}\{\theta_1\geq \theta_a\}\right)}{\mathbb P\left(A^*=a_1\right)}\\
        &=\frac{\int_{-\infty}^x\prod_{a\neq a_1}\int_{-\infty}^{\theta_1}\d\rho(\theta_a)\d \rho(\theta_1)}{\mathbb P\left(A^*=a_1\right)}\\
        &=\frac{\int_{-\infty}^x F(\theta_1)^{K-1}\d\rho(\theta_1)}{\mathbb P\left(A^*=a_1\right)}=\int_{-\infty}^x\d \mu(\theta_1)\,.
    \end{split}
\end{equation*}

Define another event $\cE_3=\{F(\theta_1)\geq 1/2\}$. Using the assumption $K\geq \log_2(T)$,
\begin{equation*}
    \mathbb P\left(\theta_1\leq 0|A^*=a_1\right) = \mathbb P\left(\max_a \theta_a\leq 0\right) = 2^{-K}\leq \frac{1}{T}\,.
\end{equation*}
Thus, we have $\mathbb P(\cE_3^c)\leq 1/T$ under the measure $\mu$.
\item  Second, we have 
\begin{equation*}
  \int_{\theta_1-\Delta}^{\theta_1}\d \rho(\theta)\leq \Delta\frac{1}{\sqrt{2\pi}}\exp\left(-\frac{(\theta_1-\Delta)^2}{2}\right)\leq \frac{\Delta}{\sqrt{2\pi}}\,.
\end{equation*}
\end{itemize}

Note that from the definition of $q(\theta_1)$ in Eq.~\eqref{def:q_theta}, $q(\theta_1)$ cannot exceed 1. Therefore, under event $\cE_3$, $q(\theta_1)$ is upper bounded by $\min(2\Delta/\sqrt{2\pi}, 1)\leq \min(\Delta, 1)$. We decompose  Eq.~\eqref{eqn:A_*} based on $\cE_3$,
\begin{equation}\label{eqn:A_bound}
 \begin{split}
    \mathbb P\left(A^*\notin\cU_{A}\right)
     &\leq
  \int_R\mathbb E\left[\left(1-\frac{\exp\left(-\beta \sqrt{2\log(TK)}/\lambda\right)}{1+K\min(\Delta, 1)+\frac{\log(T)}{\log(1/((K-1)\min(\Delta, 1)))}+K\exp(-\beta\Delta)}\right)^N\mathbb I(\cE_3) \Big|\theta_1\right]\d \mu(\theta_1)
    +\frac{3}{T}\\
     &
    \leq
  \left(1-\frac{\exp\left(-\beta \sqrt{2\log(TK)}/\lambda\right)}{1+K\min(\Delta, 1)+\frac{\log(T)}{\log(1/((K-1)\min(\Delta, 1)))}+K\exp(-\beta\Delta)}\right)^N
    +\frac{3}{T}\,.
     \end{split}
\end{equation}
With the choice of $\Delta = \log(T\beta)/\beta$, we have 
\begin{equation*}
    \begin{split}
          \mathbb P\left(A^*\notin\cU_{A}\right)
    \leq
  \left(1-\frac{\exp\left(-\beta \sqrt{2\log(TK)}/\lambda\right)}{1+K\min(\log(T\beta)/\beta, 1)+\frac{\log(T)}{\log(1/((K-1)\min(\log(T\beta)/\beta, 1)))}+K/(T\beta)}\right)^N
    +\frac{3}{T}\,.
    \end{split}
\end{equation*}
This ends the proof.

\subsection{Prove $
\mathbb E[|\cU_{A}|]\leq f_2
$}
We first observe that
\begin{equation}\label{eqn:D_0_1}
    \begin{split}
        \mathbb E[|\cU_{A}|]&=\mathbb E\left[\sum_{a\in\cA}\mathbb I(a\in \cU_{A})\right] = \mathbb E\left[\sum_{a\in\cB}\mathbb I(a\in \cU_{A})+\sum_{a\notin\cB}\mathbb I(a\in \cU_{A})\right]\\
        &\leq \mathbb E[|\cB|]+\mathbb E\left[\sum_{a\in\cA}\mathbb I\left(a\in \cU_{A}, a\notin\cB\right)\right]\,,
    \end{split}
\end{equation}
where $\cB$ is defined in Eq.~\eqref{def:set_B}.
For the second term in Eq.~\eqref{eqn:D_0_1},
\begin{equation*}
    \begin{split}
       \mathbb E\left[\sum_{a\in\cA}\mathbb I\left(a\in \cU_{A}, a\notin\cB\right)\right]&=\sum_{a\in\cA}\mathbb P\left(a\in \cU_{A}, a\notin\cB\right)\\
        &= \mathbb E\left[\sum_{a\in\cA}\mathbb P\left(a\in \cU_{A}, a\notin\cB\Big|\theta, \vartheta\right)\right]\\
        &\leq \mathbb E\left[\sum_{n=1}^N\sum_{a\in\cA}\mathbb P\left(\bar A_n=a, \langle A^*-a, \theta\rangle\geq \Delta\Big|\theta, \vartheta\right)\right]\\
        &=\mathbb E\left[\sum_{n=1}^N\sum_{a\in\cA}\mathbb P\left(\bar A_n=a, \langle A^*-a, \theta-\vartheta\rangle + \langle A^*-a, \vartheta\rangle\geq \Delta\Big|\theta, \vartheta\right)\right]\\
         &\leq \mathbb E\left[\sum_{n=1}^N\sum_{a\in\cA}\mathbb P\left(\bar A_n=a, \|A^*-a\|_{1}\| \theta-\vartheta\|_{\infty}+ \langle A^*-a, \vartheta\rangle\geq \Delta\Big|\theta, \vartheta\right)\right]\,.
    \end{split}
\end{equation*}
Similar to the proof in Appendix \ref{sec:proof_lemma1_part1}, we decompose the above based on $\cE_1$:
\begin{equation}\label{eqn:D_0_2}
    \begin{split}
         \mathbb E\left[\sum_{a\in\cA}\mathbb I\left(a\in \cU_{A}, a\notin\cB\right)\right] &= \mathbb E\left[\sum_{n=1}^N\sum_{a\in\cA}\mathbb P\left(\bar A_n=a, \langle A^*-a, \theta-\vartheta\rangle + \langle A^*-a, \vartheta\rangle\geq \Delta\Big|\theta, \vartheta\right)\mathbb I(\cE_1)\right]+\mathbb P(\cE_1^c)\\
         &\leq \mathbb E\left[\sum_{n=1}^N\sum_{a\in\cA}\mathbb P\left(\bar A_n=a, \langle A^*-a, \vartheta\rangle\geq \Delta-\sqrt{2\log(TK)}/\lambda\Big|\theta, \vartheta\right)\mathbb I(\cE_1)\right]+\frac{1}{T}\\
         &\leq N \mathbb E\left[\sum_{a\in\cA, \langle A^*-a, \vartheta\rangle\geq \Delta-\sqrt{2\log(TK)}/\lambda}\frac{\exp(\beta\langle a, \vartheta\rangle)}{\sum_{b\in\cA}\exp(\beta\langle b, \vartheta\rangle)}\right]+\frac{1}{T}\\
         &=N \mathbb E\left[\sum_{a\in\cA, \langle A^*-a, \vartheta\rangle\geq \Delta-\sqrt{2\log(TK)}/\lambda}\frac{1}{\sum_{b\in\cA}\exp(\beta\langle b-a, \vartheta\rangle)}\right]+\frac{1}{T}\\
         &\leq N \mathbb E\left[\sum_{a\in\cA, \langle A^*-a, \vartheta\rangle\geq \Delta-\sqrt{2\log(TK)}/\lambda}\frac{1}{\exp(\beta\langle A^*-a, \vartheta\rangle)}\right]+\frac{1}{T}\\
         &\leq N \mathbb E\left[\sum_{a\in\cA, \langle A^*-a, \vartheta\rangle\geq \Delta-\sqrt{2\log(TK)}/\lambda}\frac{1}{\exp(\beta(\Delta-\sqrt{2\log(TK)}/\lambda))}\right]+\frac{1}{T}\\
          &\leq NK\exp\left(-\beta(\Delta-\sqrt{2\log(TK)}/\lambda)\right)+\frac{1}{T}\,.
    \end{split}
\end{equation}
Combining Eqs.~\eqref{eqn:D_0_1}-\eqref{eqn:D_0_2} together, 
\begin{equation}\label{eqn:D_bound}
    \begin{split}
        \mathbb E[|\cU_{A}|] 
        \leq \mathbb E[|\cB|] + NK\exp\left(-\beta(\Delta-\sqrt{2\log(TK)}/\lambda)\right)+\frac{1}{T}\,.
    \end{split}
\end{equation}

Now we start to bound $\mathbb E[|\cB|]$. By the definition,
\begin{equation*}
\begin{split}
    \mathbb E\left[|\cB|-1\right] &= \sum_{a\in\cA}\mathbb E\left[|\cB|-1|A^*=a\right]\mathbb P(A^*=a)\\
    &=\sum_{a\in\cA}\sum_{k=0}^{K-1}k\mathbb P\left(|\cB|-1=k|A^*=a\right)\mathbb P(A^*=a)\,.
    \end{split}
\end{equation*}

Using Eq.~\eqref{eqn:B_prob},
\begin{equation*}
    \begin{split}
           \mathbb E\left[|\cB|-1\right] & = \sum_{a\in\cA}\sum_{k=0}^{K-1}k\int_R\mathbb P(X_{\theta_a}=k)\d \mu(\theta_a)\mathbb P(A^*=a)\\
           &=\sum_{a\in\cA}\int_R\sum_{k=0}^{K-1}k\mathbb P(X_{\theta_a}=k)\d \mu(\theta_a)\mathbb P(A^*=a)=  \sum_{a\in\cA}\int_R\mathbb E[X_{\theta_a}]\d \mu(\theta_a)\mathbb P(A^*=a)\\
           &=\sum_{a\in\cA}\int_R (K-1)q(\theta_a)\d \mu(\theta_a)\mathbb P(A^*=a)\,.
    \end{split}
\end{equation*}
Bounding $q(\theta_a)$ in a similar way, it implies 
\begin{equation*}
\begin{split}
  \mathbb E\left[|\cB|-1\right]&\leq \sum_{a\in\cA}\frac{2\Delta(K-1)}{\sqrt{2\pi}}\int_R \exp\left(-\frac{(\theta_a-\Delta)^2}{2}\right)\d \mu(\theta_a)\mathbb P(A^*=a)\\
  &\leq \frac{2}{\sqrt{2\pi}}(K-1)\Delta\leq (K-1)\Delta\,.
  \end{split}
\end{equation*}
Note that $|\cB|-1$ is at most $K-1$ such that 
\begin{equation*}
     \mathbb E\left[|\cB|-1\right]\leq (K-1)\min(\Delta, 1)\leq K\min(\Delta, 1)\,.  
\end{equation*}
Together with Eq.~\eqref{eqn:D_bound}, we can show that  
\begin{equation}\label{eqn:D_0_bound}
    \mathbb E[|\cU_{A}|]\leq K\min(\Delta, 1)+1+NK\exp\left(-\beta(\Delta-\sqrt{2\log(TK)}/\lambda)\right)+\frac{1}{T}\,.
\end{equation}
With the choice of $\Delta = \log(T\beta)/\beta$, we have 
\begin{equation*}
       \mathbb E[|\cU_{A}|]\leq K\min\left(\frac{\log(T\beta)}{\beta}, 1\right)+1+\frac{NK}{T\beta}\exp\left(\beta\sqrt{2\log(TK)}/\lambda\right)+\frac{1}{T}\,.
\end{equation*}
This ends the proof.

\section{Proof of Lemma \ref{lemma:MAP}}\label{sec:proof_map}
At time $t$, to obtain the MAP estimate, we can solve the following optimization problem:
\begin{equation*}
    \begin{split}
   \argmax_{\theta, \vartheta} \underbrace{\log  P(\mathcal{D}_{t}|\theta, \vartheta)}_{\text{log-likelihood function}}  +\underbrace{\log f(\theta, \vartheta)}_{\text{log-prior function}}\,,
    \end{split}
\end{equation*}
where the log-prior function has the form
\begin{equation}\label{eqn:prior_function}
    \begin{split}
  \log f(\theta, \vartheta) =& \log f(\vartheta|\theta)+
        \log f(\theta)\\ =&-\frac{d}{2}\log(2\pi /\lambda^2)-\frac{\lambda^2}{2}\|\vartheta-\theta\|_2^2-\frac{1}{2}\log(\det(2\pi\Sigma_0))-\frac{1}{2}\theta^{\top}\Sigma_0^{-1}\theta \,,
    \end{split}
\end{equation}
and the log-likelihood function can be defined in three steps:
\begin{itemize}
    \item First, we write this as sum of two terms, one  involving the offline dataset $\cD_0$ and the other involving the online data $\cH_t$:
\begin{equation*}
    \begin{split}
 \log  P(\mathcal{D}_{t}|\theta, \vartheta)=& \log  P\left(\cD_0|\theta, \vartheta\right)
 + \log  P\left(\cH_t|\cD_0, \theta, \vartheta\right)\,.
    \end{split}
\end{equation*}
\item Second, we decompose the log-likelihood function for the offline dataset $\cD_0$ into a sum of action likelihood and reward likelihood functions:
\begin{equation}\label{eqn:likelihood1}
    \begin{split}
 \log  P\left(\cD_0|\theta, \vartheta\right) &= \sum_{n=1}^{N}\left(\log P \left(\bar{A}_n | \theta, \vartheta \right) + \log P \left(\bar R_{n}  | \bar{A}_n, \theta, \vartheta \right) \right)\\
    &=  \sum_{n=1}^{N}\left(\beta \vartheta^T \bar{A}_n - \log\left( \sum_{b \in \mathcal{A}} \exp \left( \beta \vartheta^T b \right)\right)\right)- \frac{1}{2} \sum_{n=1}^{N} \left( \bar R_{n} - \theta^T \bar{A}_n \right)^2 -\frac{N}{2}\log(2\pi)\,,
    \end{split}
\end{equation}
where the second equation follows Eq.~\eqref{eq:softmaxactions} and Gaussian noise assumption.
\item Third, as per the TS algorithm,
$A_t$ is independent of $\theta$ conditioned on $\cU_{t-1}$, 
which implies
\begin{equation}\label{eqn:likelihood2}
\begin{split}
       \log P\left(\cH_t|\cD_0, \theta, \vartheta\right) 
       & = \sum_{\tau=1}^{t} \log P \left(A_{\tau} | \mathcal{D}_{\tau-1}\right) + \sum_{\tau=1}^{t}\log  P \left(R_{\tau}  | A_{\tau}, \theta, \vartheta \right)\\
      & = -\frac{1}{2} \sum_{\tau=1}^{t} \left( R_{\tau} - \theta^T A_{\tau} \right)^2 -\frac{t}{2}\log(2\pi) +\text{const.}
\end{split}
\end{equation}
\end{itemize}
Putting Eqs.~\eqref{eqn:prior_function}-\eqref{eqn:likelihood2} together, our loss function to obtain a MAP estimate can be simplified to
\begin{equation*}
\begin{split}
\cL(\theta, \vartheta)&= - \sum_{n=1}^{N}\left(\beta \vartheta^T \bar{A}_n - \log\left( \sum_{b \in \mathcal{A}} \exp \left( \beta \vartheta^T b \right)\right)\right)\\
& + \frac{1}{2 } \sum_{n=1}^{N} \left( \bar R_{n} - \theta^T \bar{A}_n \right)^2 +\frac{1}{2} \sum_{\tau=1}^{t} \left( R_{\tau} - \theta^T A_{\tau} \right)^2+\frac{\lambda^2}{2}\| \vartheta-\theta\|_2^2+\frac{1}{2}\theta^{\top}\Sigma_0^{-1}\theta\,.
\end{split}
\end{equation*}
This ends the proof.

\section{Proof of Claim \ref{eqn:reward_range}}\label{sec:proof_gaussian_range}
We just need to prove an upper bound for $\mathbb E[\max_k X_k]$ where $X_k\overset{\text{i.i.d}}{\sim}\cN(0, 1)$. By the Jenson's inequality,
\begin{equation*}
    \begin{split}
        \exp\left(t\mathbb E\left[\max_k X_k\right]\right)\leq \mathbb E\left[\exp\left(t\max_k X_k\right)\right] = \mathbb E\left[\max_k\exp\left(tX_k\right)\right]\leq \sum_{k=1}^K\mathbb E\left[\exp\left(tX_k\right)\right] = K\exp(t^2/2)\,,
    \end{split}
\end{equation*}
where the last equality follows from the definition of the Gaussian moment generating function. This implies 
\begin{equation*}
    \mathbb E\left[\max_k X_k\right]\leq \frac{\log(K)}{t} + \frac{t}{2}\,.
\end{equation*}
Letting $t=\sqrt{2\log(K)}$, we have 
\begin{equation*}
    \mathbb E[\max_k X_k]\leq \sqrt{2\log(K)}\,.
\end{equation*}
This ends the proof.

%% file: icml2023.bbl
\begin{thebibliography}{35}
\providecommand{\natexlab}[1]{#1}
\providecommand{\url}[1]{\texttt{#1}}
\expandafter\ifx\csname urlstyle\endcsname\relax
  \providecommand{\doi}[1]{doi: #1}\else
  \providecommand{\doi}{doi: \begingroup \urlstyle{rm}\Url}\fi

\bibitem[Banerjee et~al.(2022)Banerjee, Sinclair, Tambe, Xu, and
  Yu]{banerjee22artificial}
Banerjee, S., Sinclair, S.~R., Tambe, M., Xu, L., and Yu, C.~L.
\newblock Artificial replay: A meta-algorithm for harnessing historical data in
  bandits.
\newblock \emph{arXiv}, pp.\  2210.00025, 2022.

\bibitem[Beliaev et~al.(2022)Beliaev, Shih, Ermon, Sadigh, and
  Pedarsani]{beliaev2022imitation}
Beliaev, M., Shih, A., Ermon, S., Sadigh, D., and Pedarsani, R.
\newblock Imitation learning by estimating expertise of demonstrators.
\newblock \emph{Proceedings of the 39th International Conference on Machine
  Learning}, 162:\penalty0 1732--1748, 2022.

\bibitem[Bubeck \& Liu(2013)Bubeck and Liu]{bubeck2013prior}
Bubeck, S. and Liu, C.-Y.
\newblock Prior-free and prior-dependent regret bounds for thompson sampling.
\newblock \emph{Advances in neural information processing systems}, 26, 2013.

\bibitem[Cutkosky et~al.(2022)Cutkosky, Dann, Das, and
  Zhang]{cutkosky2022leveraging}
Cutkosky, A., Dann, C., Das, A., and Zhang, Q.
\newblock Leveraging initial hints for free in stochastic linear bandits.
\newblock In \emph{International Conference on Algorithmic Learning Theory},
  pp.\  282--318. PMLR, 2022.

\bibitem[Degenne et~al.(2018)Degenne, Garcelon, and
  Perchet]{degenne2018bandits}
Degenne, R., Garcelon, E., and Perchet, V.
\newblock Bandits with side observations: Bounded vs. logarithmic regret.
\newblock In \emph{Conference on Uncertainty in Artificial Intelligence}, 2018.

\bibitem[Diamond \& Boyd(2016)Diamond and Boyd]{diamond2016cvxpy}
Diamond, S. and Boyd, S.
\newblock {CVXPY}: {A} {P}ython-embedded modeling language for convex
  optimization.
\newblock \emph{Journal of Machine Learning Research}, 17\penalty0
  (83):\penalty0 1--5, 2016.

\bibitem[Dwaracherla et~al.(2022)Dwaracherla, Wen, Osband, Lu, Asghari, and
  Van~Roy]{dwaracherla2022ensembles}
Dwaracherla, V., Wen, Z., Osband, I., Lu, X., Asghari, S.~M., and Van~Roy, B.
\newblock Ensembles for uncertainty estimation: Benefits of prior functions and
  bootstrapping.
\newblock \emph{arXiv preprint arXiv:2206.03633}, 2022.

\bibitem[Hao \& Lattimore(2022)Hao and Lattimore]{hao2022regret}
Hao, B. and Lattimore, T.
\newblock Regret bounds for information-directed reinforcement learning.
\newblock In Oh, A.~H., Agarwal, A., Belgrave, D., and Cho, K. (eds.),
  \emph{Advances in Neural Information Processing Systems}, 2022.
\newblock URL \url{https://openreview.net/forum?id=1pHC-yZfaTK}.

\bibitem[Hao et~al.(2021)Hao, Lattimore, and Deng]{hao2021information}
Hao, B., Lattimore, T., and Deng, W.
\newblock Information directed sampling for sparse linear bandits.
\newblock \emph{Advances in Neural Information Processing Systems},
  34:\penalty0 16738--16750, 2021.

\bibitem[Kveton et~al.(2021)Kveton, Konobeev, Zaheer, Hsu, Mladenov, Boutilier,
  and Szepesvari]{kveton2021meta}
Kveton, B., Konobeev, M., Zaheer, M., Hsu, C.-w., Mladenov, M., Boutilier, C.,
  and Szepesvari, C.
\newblock Meta-thompson sampling.
\newblock In \emph{International Conference on Machine Learning}, pp.\
  5884--5893. PMLR, 2021.

\bibitem[Lattimore \& Szepesv{\'a}ri(2019)Lattimore and
  Szepesv{\'a}ri]{lattimore2019information}
Lattimore, T. and Szepesv{\'a}ri, C.
\newblock An information-theoretic approach to minimax regret in partial
  monitoring.
\newblock In \emph{Conference on Learning Theory}, pp.\  2111--2139. PMLR,
  2019.

\bibitem[Lattimore \& Szepesv{\'a}ri(2020)Lattimore and
  Szepesv{\'a}ri]{lattimore2020bandit}
Lattimore, T. and Szepesv{\'a}ri, C.
\newblock \emph{Bandit algorithms}.
\newblock Cambridge University Press, 2020.

\bibitem[Lu \& Van~Roy(2017)Lu and Van~Roy]{lu2017ensemble}
Lu, X. and Van~Roy, B.
\newblock Ensemble sampling.
\newblock \emph{Advances in neural information processing systems}, 30, 2017.

\bibitem[Osband et~al.(2018)Osband, Aslanides, and
  Cassirer]{osband2018randomized}
Osband, I., Aslanides, J., and Cassirer, A.
\newblock Randomized prior functions for deep reinforcement learning.
\newblock \emph{Advances in Neural Information Processing Systems}, 31, 2018.

\bibitem[Osband et~al.(2019)Osband, Van~Roy, Russo, Wen,
  et~al.]{osband2019deep}
Osband, I., Van~Roy, B., Russo, D.~J., Wen, Z., et~al.
\newblock Deep exploration via randomized value functions.
\newblock \emph{J. Mach. Learn. Res.}, 20\penalty0 (124):\penalty0 1--62, 2019.

\bibitem[Osband et~al.(2021)Osband, Wen, Asghari, Ibrahimi, Lu, and
  Van~Roy]{osband2021epistemic}
Osband, I., Wen, Z., Asghari, M., Ibrahimi, M., Lu, X., and Van~Roy, B.
\newblock Epistemic neural networks.
\newblock \emph{arXiv preprint arXiv:2107.08924}, 2021.

\bibitem[Osband et~al.(2022)Osband, Wen, Asghari, Dwaracherla, Lu, Ibrahimi,
  Lawson, Hao, O'Donoghue, and Roy]{osband2022the}
Osband, I., Wen, Z., Asghari, S.~M., Dwaracherla, V., Lu, X., Ibrahimi, M.,
  Lawson, D., Hao, B., O'Donoghue, B., and Roy, B.~V.
\newblock The neural testbed: Evaluating joint predictions.
\newblock In Oh, A.~H., Agarwal, A., Belgrave, D., and Cho, K. (eds.),
  \emph{Advances in Neural Information Processing Systems}, 2022.
\newblock URL \url{https://openreview.net/forum?id=JyTT03dqCFD}.

\bibitem[Ouyang et~al.(2022)Ouyang, Wu, Jiang, Almeida, Wainwright, Mishkin,
  Zhang, Agarwal, Slama, Ray, et~al.]{ouyang2022training}
Ouyang, L., Wu, J., Jiang, X., Almeida, D., Wainwright, C.~L., Mishkin, P.,
  Zhang, C., Agarwal, S., Slama, K., Ray, A., et~al.
\newblock Training language models to follow instructions with human feedback.
\newblock \emph{arXiv preprint arXiv:2203.02155}, 2022.

\bibitem[Qin et~al.(2022)Qin, Wen, Lu, and Van~Roy]{qin2022analysis}
Qin, C., Wen, Z., Lu, X., and Van~Roy, B.
\newblock An analysis of ensemble sampling.
\newblock \emph{arXiv preprint arXiv:2203.01303}, 2022.

\bibitem[Rashidinejad et~al.(2021)Rashidinejad, Zhu, Ma, Jiao, and
  Russell]{rashidinejad2021bridging}
Rashidinejad, P., Zhu, B., Ma, C., Jiao, J., and Russell, S.
\newblock Bridging offline reinforcement learning and imitation learning: A
  tale of pessimism.
\newblock \emph{Advances in Neural Information Processing Systems},
  34:\penalty0 11702--11716, 2021.

\bibitem[Ross et~al.(2011)Ross, Gordon, and Bagnell]{ross2011reduction}
Ross, S., Gordon, G., and Bagnell, D.
\newblock A reduction of imitation learning and structured prediction to
  no-regret online learning.
\newblock In \emph{Proceedings of the fourteenth international conference on
  artificial intelligence and statistics}, pp.\  627--635. JMLR Workshop and
  Conference Proceedings, 2011.

\bibitem[Russo \& Van~Roy(2014)Russo and Van~Roy]{russo2014learning}
Russo, D. and Van~Roy, B.
\newblock Learning to optimize via information-directed sampling.
\newblock \emph{Advances in Neural Information Processing Systems}, 27, 2014.

\bibitem[Russo \& Van~Roy(2016)Russo and Van~Roy]{russo2016information}
Russo, D. and Van~Roy, B.
\newblock An information-theoretic analysis of thompson sampling.
\newblock \emph{The Journal of Machine Learning Research}, 17\penalty0
  (1):\penalty0 2442--2471, 2016.

\bibitem[Russo et~al.(2018)Russo, Van~Roy, Kazerouni, Osband, Wen,
  et~al.]{russo2018tutorial}
Russo, D.~J., Van~Roy, B., Kazerouni, A., Osband, I., Wen, Z., et~al.
\newblock A tutorial on thompson sampling.
\newblock \emph{Foundations and Trends{\textregistered} in Machine Learning},
  11\penalty0 (1):\penalty0 1--96, 2018.

\bibitem[Shivaswamy \& Joachims(2012)Shivaswamy and
  Joachims]{shivaswamy2012multi}
Shivaswamy, P. and Joachims, T.
\newblock Multi-armed bandit problems with history.
\newblock In \emph{Artificial Intelligence and Statistics}, pp.\  1046--1054.
  PMLR, 2012.

\bibitem[Simchowitz et~al.(2021)Simchowitz, Tosh, Krishnamurthy, Hsu, Lykouris,
  Dudik, and Schapire]{simchowitz2021bayesian}
Simchowitz, M., Tosh, C., Krishnamurthy, A., Hsu, D.~J., Lykouris, T., Dudik,
  M., and Schapire, R.~E.
\newblock Bayesian decision-making under misspecified priors with applications
  to meta-learning.
\newblock \emph{Advances in Neural Information Processing Systems},
  34:\penalty0 26382--26394, 2021.

\bibitem[Song et~al.(2022)Song, Zhou, Sekhari, Bagnell, Krishnamurthy, and
  Sun]{song2022hybrid}
Song, Y., Zhou, Y., Sekhari, A., Bagnell, J.~A., Krishnamurthy, A., and Sun, W.
\newblock Hybrid rl: Using both offline and online data can make rl efficient.
\newblock \emph{arXiv preprint arXiv:2210.06718}, 2022.

\bibitem[Thompson(1933)]{thompson1933likelihood}
Thompson, W.~R.
\newblock On the likelihood that one unknown probability exceeds another in
  view of the evidence of two samples.
\newblock \emph{Biometrika}, 25\penalty0 (3-4):\penalty0 285--294, 1933.

\bibitem[Tossou et~al.(2017)Tossou, Dimitrakakis, and
  Dubhashi]{tossou2017thompson}
Tossou, A.~C., Dimitrakakis, C., and Dubhashi, D.
\newblock Thompson sampling for stochastic bandits with graph feedback.
\newblock In \emph{Thirty-First AAAI Conference on Artificial Intelligence},
  2017.

\bibitem[Vershynin(2010)]{vershynin2010introduction}
Vershynin, R.
\newblock Introduction to the non-asymptotic analysis of random matrices.
\newblock \emph{arXiv preprint arXiv:1011.3027}, 2010.

\bibitem[Wagenmaker \& Pacchiano(2022)Wagenmaker and
  Pacchiano]{wagenmaker2022leveraging}
Wagenmaker, A. and Pacchiano, A.
\newblock Leveraging offline data in online reinforcement learning.
\newblock \emph{arXiv preprint arXiv:2211.04974}, 2022.

\bibitem[Xie et~al.(2021)Xie, Jiang, Wang, Xiong, and Bai]{xie2021policy}
Xie, T., Jiang, N., Wang, H., Xiong, C., and Bai, Y.
\newblock Policy finetuning: Bridging sample-efficient offline and online
  reinforcement learning.
\newblock \emph{Advances in neural information processing systems},
  34:\penalty0 27395--27407, 2021.

\bibitem[Zhang et~al.(2019)Zhang, Agarwal, Daum{\'e}~III, Langford, and
  Negahban]{zhang2019warm}
Zhang, C., Agarwal, A., Daum{\'e}~III, H., Langford, J., and Negahban, S.~N.
\newblock Warm-starting contextual bandits: Robustly combining supervised and
  bandit feedback.
\newblock \emph{arXiv preprint arXiv:1901.00301}, 2019.

\bibitem[Zhang(2022)]{zhang2022feel}
Zhang, T.
\newblock Feel-good thompson sampling for contextual bandits and reinforcement
  learning.
\newblock \emph{SIAM Journal on Mathematics of Data Science}, 4\penalty0
  (2):\penalty0 834--857, 2022.

\bibitem[Ziegler et~al.(2019)Ziegler, Stiennon, Wu, Brown, Radford, Amodei,
  Christiano, and Irving]{ziegler2019fine}
Ziegler, D.~M., Stiennon, N., Wu, J., Brown, T.~B., Radford, A., Amodei, D.,
  Christiano, P., and Irving, G.
\newblock Fine-tuning language models from human preferences.
\newblock \emph{arXiv preprint arXiv:1909.08593}, 2019.

\end{thebibliography}
